\documentclass[10pt,twocolumn,letterpaper]{article}

\usepackage[pagenumbers]{cvpr} 

\usepackage{graphicx}
\usepackage{amsmath}
\usepackage{amssymb}
\usepackage{booktabs}

\usepackage{color}
\usepackage{soul}
\usepackage{xcolor}
\usepackage{multirow}
\usepackage{booktabs}
\usepackage{enumerate}
\usepackage{enumitem}
\usepackage{pifont}%

\usepackage{color, colortbl}
\usepackage{multirow}
\usepackage{wrapfig}
\usepackage{soul}
\definecolor{lightgray}{rgb}{0.83, 0.83, 0.83}
\definecolor{aliceblue}{rgb}{0.94, 0.97, 1.0}
\definecolor{rose}{rgb}{0.84, 0.09, 0.41}

\usepackage[pagebackref,breaklinks,colorlinks]{hyperref}

\usepackage[capitalize]{cleveref}
\crefname{section}{Sec.}{Secs.}
\Crefname{section}{Section}{Sections}
\Crefname{table}{Table}{Tables}
\crefname{table}{Tab.}{Tabs.}

\def\cvprPaperID{12522} 
\def\confName{CVPR}
\def\confYear{2023}

\begin{document}

\title{FeatER: An Efficient Network for Human Reconstruction via \underline{Feat}ure Map-Based Transform\underline{ER} }

\author{Ce Zheng$^{1}$, Matias Mendieta$^{1}$, Taojiannan Yang$^{1}$, Guo-Jun Qi$^{2,3}$,  Chen Chen$^{1}$\\
$^1$Center for Research in Computer Vision, University of Central Florida\\
$^2$  OPPO Seattle Research Center, USA \quad $^{3}$ Westlake University\\
{\tt\small \{cezheng,mendieta,taoyang1122\}@knights.ucf.edu;  guojunq@gmail.com; chen.chen@crcv.ucf.edu}\\
}


\maketitle


\begin{abstract}
Recently, vision transformers have shown great success in a set of human reconstruction tasks such as 2D/3D human pose estimation (2D/3D HPE) and human mesh reconstruction (HMR) tasks. In these tasks, feature map representations of the human structural information are often extracted first from the image by a CNN (such as HRNet), and then further processed by transformer to predict the heatmaps 
for HPE or HMR. However, existing transformer architectures are not able to process these feature map inputs directly, forcing an unnatural flattening of the location-sensitive human structural information. Furthermore, much of the performance benefit in recent HPE and HMR methods has come at the cost of ever-increasing computation and memory needs. Therefore, to simultaneously address these problems, we propose FeatER, a novel transformer design that preserves the inherent structure of feature map representations when modeling attention while reducing memory and computational costs. Taking advantage of FeatER, we build an efficient network for a set of human reconstruction tasks including 2D HPE, 3D HPE, and HMR. A feature map reconstruction module is applied to improve the performance of the estimated human pose and mesh. Extensive experiments demonstrate the effectiveness of FeatER on various human pose and mesh datasets. For instance, FeatER outperforms the SOTA method MeshGraphormer by requiring 5\% of Params and 16\% of MACs on Human3.6M and 3DPW datasets.  \textcolor{magenta}{The project
webpage is \url{https://zczcwh.github.io/feater_page/}}.
\end{abstract}

\section{Introduction}

Understanding human structure from monocular images is one of the fundamental topics in computer vision. The corresponding tasks of Human Pose Estimation (HPE) and Human Mesh Reconstruction (HMR) have received a growing interest from researchers, accelerating progress toward various applications such as VR/AR, virtual try-on, and AI coaching. However, HPE and HMR from a single image still remain challenging tasks due to depth ambiguity, occlusion, and complex human body articulation.

\begin{figure}[htp]
\vspace{-5pt}
  \centering
  \includegraphics[width=0.80\linewidth]{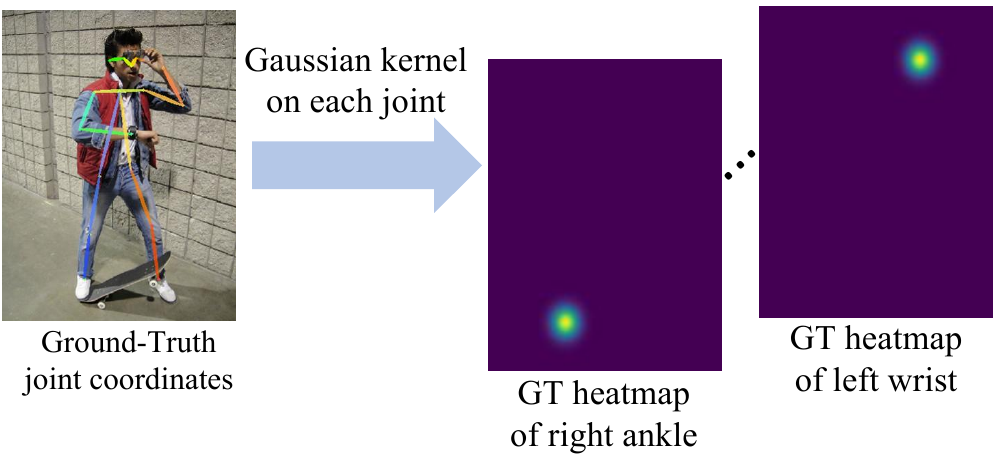}
    \vspace{-5pt}
  \caption{Generating heatmaps from joint coordinates. }
  \vspace{-5pt}
  \label{fig:heatmap_rep}
\end{figure}

With the blooming of deep learning techniques, Convolutional Neural Network (CNN) \cite{vgg,resnet,hrnet} architectures have been extensively utilized in vision tasks and have achieved impressive performance. Most existing HPE and HMR models \cite{hrnet,OCHMR} utilize CNN-based architectures (such as ResNet \cite{resnet} and HRNet \cite{hrnet}) to predict feature maps, which are supervised by the ground-truth 2D heatmap representation (encodes the position of each keypoint into a feature map with a Gaussian distribution) as shown in Fig. ~\ref{fig:heatmap_rep}.
This form of output representation and supervision can make the training process smoother, and therefore has become the \textit{de facto} process in HPE's networks \cite{openpose,hrnet,darkpose_2020}.

Recently, the transformer architecture has been fruitfully adapted from the field of natural language processing (NLP) into computer vision, where it has enabled state-of-the-art performance in HPE and HMR tasks \cite{transpose2021,tokenpose2021,Poseformer_2021_ICCV,lin2021metro,lin2021_mesh_graphormer}.
The transformer architecture demonstrates a strong ability to model global dependencies in comparison to CNNs via its self-attention mechanism. The long-range correlations between tokens can be captured, which is critical for modeling the dependencies of different human body parts in HPE and HMR tasks. 
Since feature maps concentrate on certain human body parts, we aim to utilize the transformer architecture to refine the coarse feature maps (extracted by a CNN backbone).
After capturing the global correlations between human body parts, more accurate pose and mesh can be obtained.

However, inheriting from NLP where transformers embed each word to a feature vector, Vision Transformer architectures such as ViT \cite{Dosovitskiy2020ViT} can only deal with the flattened features when modeling attention. This is less than ideal for preserving the structural context of the feature maps during the refinement stage (feature maps with the shape of $[n,h,w]$ need to be flattened as $[n, d]$, where $d = h \times w $. Here $n$ is the number of feature maps, $h$ and $w$ are height and width of each feature map, respectively). Furthermore, another issue is that the large embedding dimension caused by the flattening process makes the transformer computationally expensive. This is not  suitable for real-world applications of HPE and HMR, which often demand real-time processing capabilities on deployed devices (e.g. AR/VR headsets).

Therefore, we propose a Feature map-based transformER (FeatER) architecture to properly refine the coarse feature maps through global correlations of structural information in a resource-friendly manner. Compared to the vanilla transformer architecture, FeatER has two advantages:   

\begin{itemize}[leftmargin=*]
\item First, FeatER preserves the feature map representation in the transformer encoder when modeling self-attention, which is naturally adherent with the HPE and HMR tasks. Rather than conducting the self-attention based on flattened features, FeatER ensures that the self-attention is conducted based on the original 2D feature maps, which are more structurally meaningful. 
To accomplish this, FeatER is designed with a novel dimensional decomposition strategy to handle the extracted stack of 2D feature maps.
\item Second, this decompositional design simultaneously provides a significant reduction in computational cost compared with the vanilla transformer 
\footnote{For example, there are 32 feature maps with overall dimension $[32,64,64]$. For a vanilla transformer, without discarding information, the feature maps need to be flattened into $[32, 4096]$. One vanilla transformer block requires 4.3G MACs. Even if we reduce the input size to $[32, 1024]$, it still requires 0.27G MACs. However, given the original input of $[32,64,64]$, FeatER only requires 0.09G MACs.}. This makes FeatER more suitable for the needs of real-world applications.

\end{itemize}

Equipped with FeatER, we present an {efficient} framework for human representation tasks including 2D HPE, 3D HPE, and HMR. For the more challenging 3D HPE and HMR portion, a feature map reconstruction module is integrated into the framework. Here, a subset of feature maps are randomly masked and then reconstructed by FeatER, enabling more robust 3D pose and mesh predictions for in-the-wild inference. We conduct extensive experiments on human representation tasks, including 2D human pose estimation on COCO, 3D human pose estimation and human mesh reconstruction on Human3.6M and 3DPW datasets. Our method (FeatER) consistently outperforms SOTA methods on these tasks with significant computation and memory cost reduction (e.g. FeatER outperforms MeshGraphormer \cite{lin2021_mesh_graphormer} with only requiring 5\% of Params and 16\% of MACs).

\section{Related work}
Since Vision Transformer (ViT) \cite{Dosovitskiy2020ViT} introduced the transformer architecture to image classification with great success, it has also been shown to have enormous potential in various vision tasks such as object detection \cite{misra2021end,liu2021group}, facial expression recognition
\cite{xue2021transfer}, and re-identification \cite{li2021reid,TransReID}. Since the related work of HPE and HMR is vast, we refer interested readers to the recent and comprehensive surveys: \cite{hpesurvey_chen} for HPE and  \cite{hmrsurvey} for HMR. In this section, we discuss the more relevant transformer-based approaches.

\textbf{Transformers in HPE:} HPE can be categorized into 2D HPE and 3D HPE based on 2D pose output or 3D pose output. Recently, several methods \cite{transpose2021,tokenpose2021,PRTR} utilize transformers in 2D HPE. TransPose \cite{transpose2021} uses a transformer to capture the spatial relationships between keypoint joints. PRTR \cite{PRTR} builds cascade transformers with encoders and decoders based on DETR \cite{detr}. Although achieving impressive performance, TransPose \cite{transpose2021} and PRTR \cite{PRTR} suffer from heavy computational costs.  HRFormer \cite{hrformer} integrates transformer blocks in the HRNet structure to output 2D human pose. TokenPose \cite{tokenpose2021} embeds each keypoint to a token for learning constraint relationships by transformers, but it is limited to the 2D HPE task. 
At the same time, PoseFormer \cite{Poseformer_2021_ICCV} and Li et al. \cite{li2022exploiting} first apply transformers in 3D HPE. MHFormer \cite{li2021mhformer} generates multiple plausible pose hypotheses using transformers to lift 2D pose input to 3D pose output. As 2D-3D lifting approaches, these methods \cite{Poseformer_2021_ICCV,li2022exploiting,li2021mhformer} rely on the external 2D pose detector, which is not end-to-end.
In contrast, our FeatER is an end-to-end network for the 2D HPE, 3D HPE, and HMR in a resource-friendly manner. 

\begin{figure*}[htp]
\vspace{-10pt}
  \centering
  \includegraphics[width=0.94\linewidth]{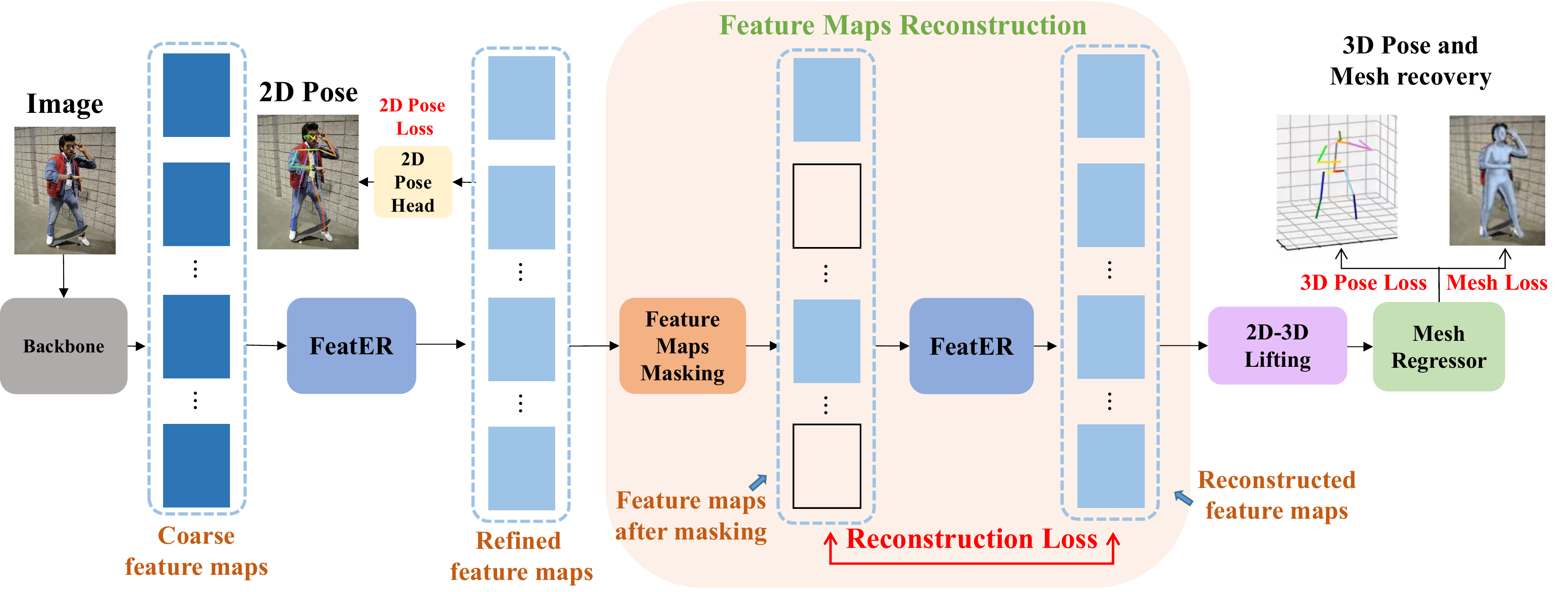}
  \caption{An overview of our proposed network for 2D HPE, 3D HPE, and HMR tasks. The coarse feature maps are extracted by the CNN backbone and refined by FeatER blocks. The 2D pose can be obtained by a 2D pose head. Then, we apply a feature map reconstruction module to improve the robustness of the predicted 3D pose and mesh. This is accomplished by randomly masking out some feature maps and utilizing FeatER blocks to reconstruct them. Next, we apply a 2D-3D lifting module which converts the 2D feature maps to 3D feature maps, and predicts the parameters for the mesh regressor. Finally, the mesh regressor outputs the 3D human pose and mesh.} 
  \label{fig:framework}
  \vspace{-5pt}
\end{figure*}

\textbf{Transformers in HMR:} THUNDR \cite{Thundr} introduces a model-free transformer-based architecture for human mesh reconstruction. 
GTRS \cite{gtrs} proposes a graph transformer architecture with parallel design to estimate 3D human mesh only from detected 2D human pose. METRO \cite{lin2021metro} combines a CNN backbone with a transformer network to regress human mesh vertices directly for HMR. MeshGraphormer \cite{lin2021_mesh_graphormer} further injects GCNs into the transformer encoder in METRO to improve the interactions among neighboring joints. Despite their excellent performance, METRO and MeshGraphormer still incur substantial memory and computational overhead.    


\textbf{Efficient Methods for HPE and HMR:} The SOTA methods for HPE and HMR \cite{lin2021metro,lin2021_mesh_graphormer} mainly pursue higher accuracy without considering computation and memory cost. While less studied, model efficiency is also a key characteristic of HPE and HMR applications. Lite-HRNet \cite{LiteHRNet} applies the efficient shuffle block in ShuffleNet \cite{shufflenetv1, shufflenetv2} to HRNet \cite{hrnet}, but it is only limited to 2D HPE. GTRS \cite{gtrs} is a lightweight pose-based method that can reconstruct human mesh from 2D human pose. However, to reduce the computation and memory cost, GTRS only uses 2D pose as input and therefore misses some information such as human shape. Thus, the performance is not comparable to the SOTA HMR methods \cite{dsr2021,lin2021_mesh_graphormer}. Our FeatER is an efficient network that can outperform SOTA methods while reducing computation and memory costs significantly.

\section{Methodology}
\subsection{Overview Architecture}

As shown in Fig. \ref{fig:framework}, we propose a network for 2D HPE, 3D HPE, and HMR tasks. Given an image, a CNN backbone is applied first to extract the coarse feature maps. Then, our proposed FeatER blocks further refine the feature maps by capturing the global correlations between them. Next, a 2D pose head is used to output the 2D pose. To improve the robustness of the estimated 3D pose and mesh, we apply a feature map reconstruction module with a masking strategy. Specifically, a subset of the feature maps are randomly masked with a fixed probability, and then FeatER blocks are tasked with reconstruction. Finally, a 2D-3D lifting module and a mesh regressor HybrIK \cite{hybrik} output the estimated 3D pose and mesh.

\subsection{Preliminaries of Vanilla Transformer}
\label{sec:vanillatrans}


The input of a vanilla transformer \cite{Dosovitskiy2020ViT} block is  $ X_{in} \in \mathbb{R} ^{n \times d} $, where $n$ is the number of patches and $d$ is the embedding dimension. Vanilla transformer block is composed of the following operations, and is applied to capture global dependencies between patches via self-attention mechanism. 

Multi-head Self-Attention Layer (MSA) is the core function to achieve self-attention modeling. After layer normalization,  the input $X_{in} \in \mathbb{R} ^{n \times d} $ is first mapped to three matrices: query matrix $Q$, key matrix $K$ and value matrix $V$ by three linear transformation: 
\begin{align}
\small
    Q = {X_{in}}W_Q, \quad K = {X_{in}}W_K, \quad V = {X_{in}}W_V.
\end{align}
where $W_Q$, $W_K$ and $W_V$ $\in \mathbb{R} ^{d \times d}$.

The scaled dot product attention can be described as the following mapping function: 
\begin{align}
\small
    {\rm Attention}(Q,K,V) = {\rm Softmax}(QK^\top/ \sqrt{d})V.
\end{align}
where $\frac{1}{\sqrt{d}}$ is the scaling factor for appropriate normalization to prevent extremely small gradients.

\begin{figure*}[htp]
\vspace{-5pt}
  \centering
  \includegraphics[width=0.91\linewidth]{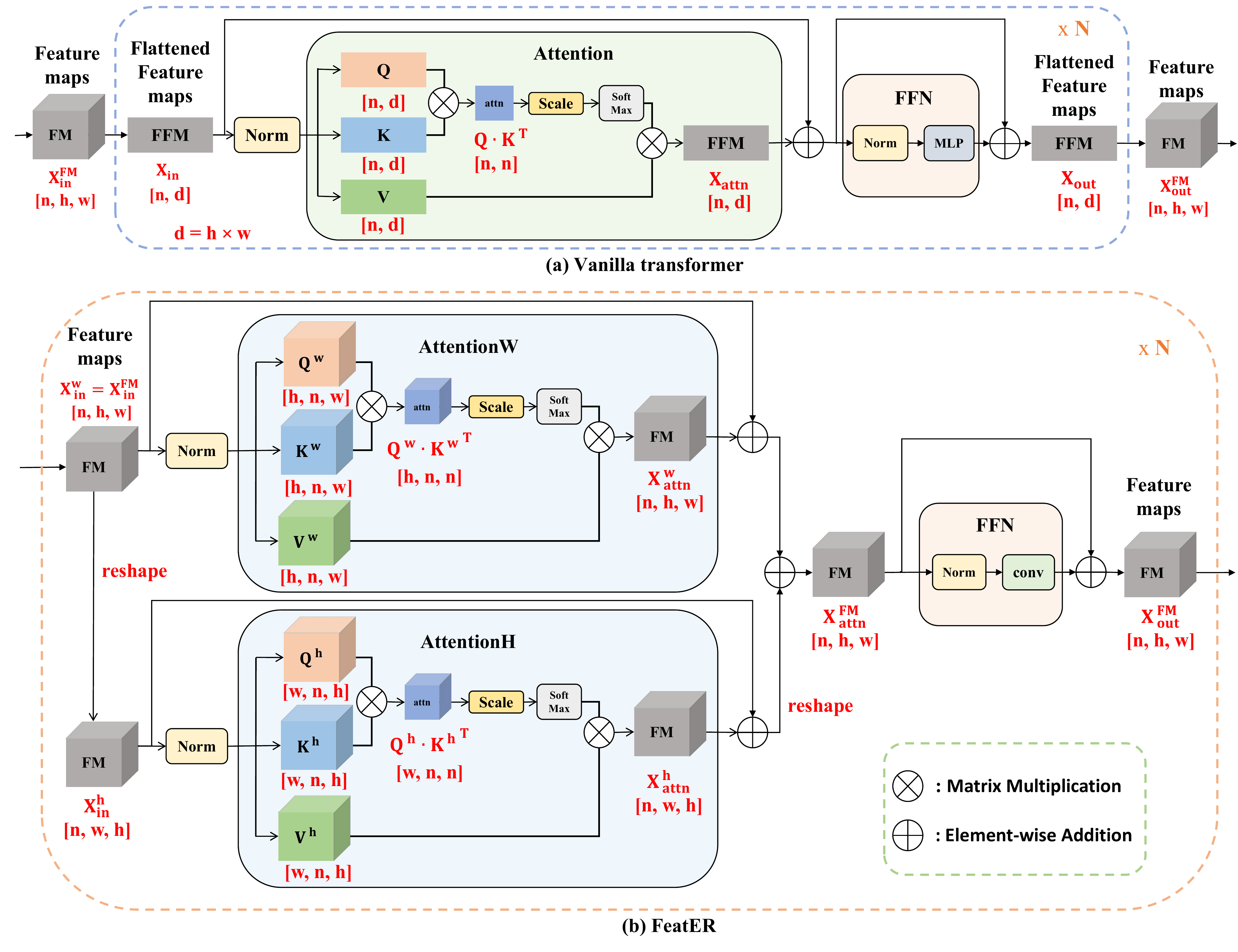}
  \vspace{-10pt}
  \caption{(a) The vanilla transformer blocks to process feature maps. (b) Our proposed FeatER blocks to process feature maps.}
  \label{fig:encoder}
  \vspace{-10pt}
\end{figure*}

Next, the vanilla transformer block architecture consisting of MSA and feed-forward network (FFN) is shown in  Fig.~\ref{fig:encoder} (a). The block output $X_{out} \in \mathbb{R} ^{n \times d}$ keeps the same size as the block input $X_{in} \in \mathbb{R} ^{n \times d}$, and is represented as follows:
\begin{flalign}
\small
   & X_{attn} = {\rm MSA}(Q,K,V) + X_{in} \\
   & X_{out} = {\rm FFN}(X_{attn}) + X_{attn}
\end{flalign}

\noindent where  $\rm MSA(\cdot)$ represents the Multi-head Self-Attention block, and $\rm FFN(\cdot)$ is a feed-forward network consisting of the multilayer perceptron (MLP) and normalization layer.

Thus, given a sequence of coarse 2D feature maps (FM) extracted by the CNN backbone $X^{FM}_{in} \in \mathbb{R} ^{n \times h \times w}$, the vanilla transformer block needs to flatten the feature maps into $X_{in} \in \mathbb{R} ^{n \times d}$, where $d = h \times w$. After the vanilla transformer block, the output $X_{out} \in \mathbb{R} ^{n \times d}$ should be converted back to feature map representation $X^{FM}_{out} \in \mathbb{R} ^{n \times h \times w}$,
which is unnatural.
Also, the large $d$ makes the transformer blocks computationally expensive.

\subsection{FeatER}

The purpose of applying transformer is to model the global correlations between a sequence of feature maps that corresponding to different human body parts. We want to preserve the inherent structure of 2D feature map representation when modeling self-attention in transformer blocks. However, as mentioned in the above section, the vanilla transformer is not able to model the self-attention given a sequence of feature maps input $X^{FM}_{in} \in \mathbb{R} ^{n \times h \times w}$. All feature maps have to be flattened into $X_{in} \in \mathbb{R} ^{n \times d}$ before the transformer blocks. The output flattened feature maps also need to be converted back to form the feature map representation $X^{FM}_{out} \in \mathbb{R} ^{n \times h \times w}$.  

\textit{Is there a better transformer architecture to deal with feature map inputs and return the feature map outputs \ul{directly and effectively?}} Motivated by this question, we propose a new Feature map-based transformER (FeatER) architecture that preserves the feature map representation when modeling self-attention for HPE and HMR tasks.

FeatER can be treated as the decomposition along $h$ and $w$ dimension of the vanilla transformer, which is illustrated in  Fig.~\ref{fig:encoder} (b). 
For $w$ dimension stream MSA (w-MSA), the input $X_{in}^{w} \in \mathbb{R} ^{n \times h \times w} $ (equals to $X^{FM}_{in} \in \mathbb{R} ^{n \times h \times w}$) is first mapped to three matrices: query matrix $Q^w$, key matrix $K^w$ and value matrix $V^w$ by three linear transformation: 
\begin{align}
\small
    Q^w = {X_{in}^w}W_{Q^w}, \quad K^w = {X_{in}^w}W_{K^w}, \quad V^w = {X_{in}^w}W_{V^w}.
\end{align}
where $W_{Q^w}$, $W_{K^w}$ and $W_{V^w}$ $\in \mathbb{R} ^{w \times w}$.

The scaled dot product attention can be described as the following mapping function: 
\begin{align}
\small
    {\rm AttentionW}(Q^w,K^w,V^w) = {\rm Softmax}(Q^w{K^w}^\top/ \sqrt{w})V^w.
\end{align}

For $h$ dimension stream MSA (h-MSA), the input $X^{FM} \in \mathbb{R} ^{n \times h \times w}$ is reshape to $X_{in}^{h}  \in \mathbb{R} ^{n \times w \times h} $, then mapped to three matrices: query matrix $Q^h$, key matrix $K^h$ and value matrix $V^h$ by three linear transformation: 
\begin{align}
\small
    Q^h = {X_{in}^{h} }W_{Q^h}, \quad K^w = {X_{in}^{h} }W_{K^h}, \quad V^w = {X_{in}^{h} }W_{V^h}.
\end{align}
where $W_{Q^h}$, $W_{K^h}$ and $W_{V^h}$ $\in \mathbb{R} ^{h \times h}$.

The scaled dot product attention can be described as the following mapping function: 
\begin{align}
\small
    {\rm AttentionH}(Q^h,K^h,V^h) = {\rm Softmax}(Q^h {K^h}^\top/ \sqrt{h})V^h.
\end{align}

Then, the FeatER block consisting of w-MSA (multi-head AttentionW), h-MSA (multi-head AttentionH) and FFN with a layer normalization operator is shown in  Fig.~\ref{fig:encoder} (b). The block output $ X_{out}^{FM} \in \mathbb{R} ^{n \times h \times w} $ keeps the same size as the input $ X_{in}^{FM} \in \mathbb{R} ^{n \times h \times w} $, and is represented as follows:
\begin{flalign}
\small
   & {X}_{attn}^{FM} = {\textrm{w-MSA}}(Q^w,K^w,V^w) + {\textrm{h-MSA}}(Q^h,K^h,V^h)^*  \\
   & {X}_{attn}^{FM} = {X}_{attn}^{FM}  +  X_{in}^{FM} \\
   & X_{out}^{FM} = {\rm FFN}( {X}_{attn}^{FM} ) +  {X}_{attn}^{FM} 
\end{flalign}

\noindent where  * means to reshape the matrix to the proper shape (i.e. from $ \mathbb{R} ^{n \times w \times h}$ to $\mathbb{R} ^{n \times h \times w} $). The $\rm FFN(\cdot)$ denotes the feed-forward network to process feature map-size input (details in \textcolor{blue}{Supplementary \ref{sec:supp_Complexity}}).  

\textbf{Complexity}: A further benefit of our FeatER design is that it inherently reduces the operational computation. The theoretical computational complexity $\Omega$ of one vanilla transformer block and one FeatER block can be approximately estimated as: 
\begin{flalign}
\small
    \Omega(\textrm{vanilla \ transformer}) &= 8nd^2 + 2n^2d \\
    \Omega(\textrm{FeatER}) &= 3nhw(w+h) + 9n^2hw
\end{flalign}
If $d = h \times w$ and $h=w$, the computational complexity of FeatER can be rewritten as $\Omega(\textrm{FeatER}) = 6nd(\sqrt{d}) + 9n^2d$. 
Normally $d$ is much larger than $n$, which means that the first term consumes the majority of the computational resource. The detailed computation comparison between the vanilla transformer block and the FeatER block is provided in  \textcolor{blue}{Supplementary \ref{sec:supp_Complexity}}. Thus, FeatER reduces the computational complexity from $\mathcal{O}(d^2)$ to $\mathcal{O}(d^{3/2})$.

\subsection{Feature Map Reconstruction Module}
Compared with estimating 2D human pose, recovering 3D pose and mesh are more challenging due to occlusion. Some joints may be occluded by the human body or other objects in the image.
In order to improve the generalization ability of our network, we apply the masking and reconstruction strategy to make our predicted human mesh more robust. 
Given a stack of refined feature maps $X^{FM} \in \mathbb{R} ^{n \times h \times w}$, we randomly mask out $m$ feature maps from $n$ feature maps (the masking ratio is $m/n$) and utilize FeatER blocks to reconstruct feature maps. The reconstruction loss computes the mean squared error (MSE) between the reconstructed and original stack of feature maps. Then, the reconstructed stack of feature maps is used for recovering 3D pose and mesh. For the inference, the feature map masking procedure is not applied. Here we only apply the feature map reconstruction module for the 3D part. More detailed discussion is provided in Section \ref{sec:Ablation}.   


\section{Experiments}
\subsection{Implementation Details}
\label{sec:Implementationdetails}
We implemented our method FeatER for HPE and HMR tasks with Pytorch \cite{PyTorch}, which is trained on four NVIDIA RTX A5000 GPUs. We first train FeatER for the 2D HPE task. The employed CNN backbone is a portion of HRNet-w32 \cite{hrnet} (first 3 stages) pretrained on COCO \cite{lin2014mscoco} dataset. There are 8 FeatER blocks for modeling global dependencies across all feature maps. The 2D pose head is a convolution block to output heatmaps of all joints. Then, we load those pretrained weights to further train the entire pipeline as illustrated in Fig. \ref{fig:framework} for 3D HPE and HMR tasks. There are another 8 FeatER blocks to recover feature maps in the feature map reconstruction module, where the masking ratio is 0.3. Then, we apply a 2D-3D lifting module to lift the 2D feature maps to 3D feature maps, and predict the parameters needed for the regressor. More details are provided in \textcolor{blue}{Supplementary \ref{sec:supp_Lifting} and \ref{sec:supp_loss}}. Next, we adopt the mesh regressor HybrIK \cite{hybrik} to output the final 3D pose and mesh. We use Adam \cite{kingma2014adam} optimizer with a learning rate of $2 \times 10^{-4} $. The batch size is 24 for each GPU.     

\begin{table*}[htp]
\renewcommand\arraystretch{1.1}
\centering
  \caption{2D Human Pose Estimation performance comparison with SOTA methods on the COCO validation set. The reported Params and MACs of FeatER are computed from the entire pipeline.}
  \vspace{2pt}
  \resizebox{0.98\linewidth}{!}
  {
  \begin{tabular}{ccccccccccc}
\hline
Model            & \multicolumn{1}{c|}{Year}                           & Input size & Params (M) & \multicolumn{1}{c|}{MACs (G)} & AP $\uparrow$           & AP50 $\uparrow$         & AP75 $\uparrow$         & AP(M) $\uparrow$        & AP(L) $\uparrow$         & AR $\uparrow$           \\ \hline
\multicolumn{2}{c}{Compared with Small Networks}                       &             &            &                                &               &               &               &               &               &               \\ \hline
DY-MobileNetV2 ~\cite{chen2020dynamic}   & \multicolumn{1}{c|}{CVPR 2020}                      & 256$\times$192     & 16.1       & \multicolumn{1}{c|}{1.0}       & 68.2          & 88.4          & 76.0          & 65.0          & 74.7          & 74.2          \\
HRFormer\_S  ~\cite{hrformer}     & \multicolumn{1}{c|}{NeurIPS 2021}                   & 256$\times$192     & 7.8        & \multicolumn{1}{c|}{2.8}       & 74.0          & \textbf{90.2}          & 81.2          & 70.4          & 80.7          & 79.4          \\
Transpose\_H\_S ~\cite{transpose2021}  & \multicolumn{1}{c|}{ICCV 2021}                      & 256$\times$192     & 8.0        & \multicolumn{1}{c|}{10.2}      & 74.2          & -             & -             & -             & -             & 78.0          \\
Tokenpose\_B  ~\cite{tokenpose2021}    & \multicolumn{1}{c|}{ICCV 2021}                      & 256$\times$192     & 13.5       & \multicolumn{1}{c|}{5.7}       & 74.7          & 89.8          & 81.4          & \textbf{71.3}          & 81.4 & \textbf{80.0}          \\
\rowcolor{lightgray}
FeatER             & \multicolumn{1}{c|}{}                               & 256$\times$192     & 8.1        & \multicolumn{1}{c|}{5.4}       & \textbf{74.9} & 89.8 & \textbf{81.6}          & 71.2 & \textbf{81.7}          & \textbf{80.0}           \\ \hline
\multicolumn{2}{c}{Compared with Large Networks}                       &             &            &                                &               &               &               &               &               &               \\ \hline
SimpleBaseline ~\cite{xiao2018simple}  & \multicolumn{1}{c|}{ECCV 2018}                      & 256$\times$192     & 34.0       & \multicolumn{1}{c|}{8.9}       & 70.4          & 88.6          & 78.3          & -             & -             & 76.3          \\
HRNet\_W32  ~\cite{hrnet}     & \multicolumn{1}{c|}{CVPR 2019}                      & 256$\times$192     & 28.5       & \multicolumn{1}{c|}{7.1}       & 74.4          & \textbf{90.5}          & \textbf{81.9}          & -             & -             & 78.9          \\
PRTR     ~\cite{PRTR}        & \multicolumn{1}{c|}{CVPR 2021}                      & 384$\times$288     & 57.2       & \multicolumn{1}{c|}{21.6}      & 73.1          & 89.4          & 79.8          & 68.8          & 80.4          & 79.8          \\
PRTR     ~\cite{PRTR}        & \multicolumn{1}{c|}{CVPR 2021}                      & 512$\times$384     & 57.2       & \multicolumn{1}{c|}{37.8}      & 73.3          & 89.2          & 79.9          & 69.0          & 80.9          & \textbf{80.2}          \\
\rowcolor{lightgray}
FeatER             & \multicolumn{1}{c|}{}                               & 256$\times$192     & 8.1        & \multicolumn{1}{c|}{5.4}       & \textbf{74.9} & 89.8 & 81.6          & \textbf{71.2} & \textbf{81.7}          & 80.0          \\ \hline
\end{tabular}
}
\label{tab: 2dhpe}
\vspace{-10pt}
\end{table*}

\subsection{2D HPE}
\label{sec:main_2DHPE}
\textbf{Dataset and evaluation metrics:} We conduct the 2D HPE experiment on the COCO \cite{lin2014mscoco} dataset, which contains over 200,000 images and 250,000 person instances. There are 17 keypoints labeled for each person instance. We train our model on the COCO train2017 set and evaluate on the COCOval2017 set, with the experiment setting following \cite{tokenpose2021}. The evaluation metrics we adopt are standard Average Precision (AP) and Average Recall (AR) \cite{hpesurvey_chen}. 

\textbf{Results:} 
Table \ref{tab: 2dhpe} compares FeatER with previous SOTA methods for 2D HPE on COCO validation set including the total parameters (Params) and Multiply–Accumulate operations (MACs). Since FeatER is a lightweight model, we first compare with previous lightweight methods (Params $\leq$ 20M and MACs $\leq$ 20G) with the input image size of 256 $\times$ 192. Our FeatER achieves the best results on 4 (AP, AP75, AP(L), and AR) of the 6 evaluation metrics. When compared with the SOTA lightweight transformer-based method Tokenpose\_B \cite{tokenpose2021}, FeatER only requires 60\% of Params and 95\% of MACs while improving 0.2 AP, 0.2 AP75, and 0.3 AP(L).

As an efficient lightweight model, FeatER can even achieve competitive performance with methods of large models while showing a significant reduction in Params and MACs. For instance, FeatER outperforms PRTR \cite{hrnet} in terms of AP, AP50, AP75, AP(M) and AP(L) with only 14\% of Params and 14\% of MACs. 

\subsection{3D HPE and HMR}
\label{sec:main_3DMesh}
\textbf{Datasets and evaluation metrics:}
We evaluate FeatER for 3D HPE and HMR on Human3.6M \cite{h36m_pami} and 3DPW \cite{pw3d2018} datasets. Human3.6M is one of the largest indoor datasets which contains 3.6M video frames in 17 actions performed by 11 actors. Following previous work \cite{Kolotouros2019SPIN,Choi_2020_ECCV_Pose2Mesh,lin2021metro}, we use 5 subjects (S1, S5, S6, S7, S8) for training and 2 subjects (S9, S11) for testing. 3DPW is an in-the-wild dataset that is composed of 60 video sequences (51K frames). The accurate 3D mesh annotations are provided. We follow the standard train/test split from the dataset. Mean Per Joint Position Error (MPJPE) \cite{hpesurvey_chen} and the MPJPE after Procrustes alignment (PA-MPJPE) \cite{hpesurvey_chen} are reported on Human3.6M. Besides these, to evaluate the reconstructed mesh, the Mean Per Vertex Error (MPVE) is reported on 3DPW. Following \cite{Kolotouros2019SPIN,hybrik,Moon_I2L_MeshNet}, Human3.6M, MPI-INF-3DHP, COCO, 3DPW are used for mixed training. Following previous work \cite{Kolotouros2019SPIN}\cite{lin2021metro}\cite{lin2021_mesh_graphormer}, the 3D human pose is calculated from the estimated mesh multiplied with the defined joint regression matrix.

\textbf{Results:} 
Table \ref{tab: 3dmesh} compares FeatER with previous SOTA methods for 3D HPE and HMR on Human3.6M and 3DPW including the Params and MACs. 
FeatER outperforms the SOTA methods on Human3.6M and 3DPW datasets with very low Params and MACs. For Human3.6M dataset, 
FeatER reduces 1.3 of MPJPE and 1.7 of PA-MPJPE compared with SOTA method MeshGraphormer \cite{lin2021_mesh_graphormer}.

For 3DPW, FeatER improves the MPJPE from 89.7 \cite{OCHMR} to 88.4, the PA-MPJPE from 55.8 \cite{TCMR_Choi_2021} to 54.5, and MPVE from 107.1 \cite{OCHMR} to 105.6 without using 3DPW training set. 
When using 3DPW training set during training, FeatER also shows superior performance compared to MeshGraphormer \cite{lin2021_mesh_graphormer}. 
Moreover, FeatER reduces the memory and computational costs significantly (only 5\% of Params and 16\% of MACs compared with MeshGraphormer \cite{lin2021_mesh_graphormer}). \ul{Thus, FeatER is a much more time and resource-efficient model for HPE and HMR tasks with exceptional performance.}

\begin{table*}[htp]
\vspace{-5pt}
\centering
  \caption{3D Pose and Mesh performance comparison with SOTA methods on Human3.6M and 3DPW datasets. The reported Params and MACs of FeatER are computed from the entire pipeline. $\dagger$ indicates video-based methods. The result of HybrIK* is with predicted camera parameters and ResNet34 is used as the backbone.}
  \vspace{-5pt}
  \resizebox{0.98\linewidth}{!}
  {
  \begin{tabular}{ll|cc|cc|ccc}
\hline
                & \multicolumn{1}{c|}{}     &               &              & \multicolumn{2}{c|}{Human3.6M} & \multicolumn{3}{c}{3DPW}                       \\ \hline
Model           & \multicolumn{1}{c|}{Year} & Params (M)     & MACs (G)      & MPJPE$\downarrow$          & PA-MPJPE$\downarrow$      & MPJPE$\downarrow$         & PA-MPJPE$\downarrow$      & MPVE$\downarrow$           \\ \hline
SPIN     ~\cite{Kolotouros2019SPIN}        & ICCV 2019                 & -             & -            & 62.5           & 41.1          & 96.9          & 59.2          & 116.4          \\
VIBE $\dagger$    ~\cite{kocabas2020vibe}         & CVPR 2020                 & -             & -            & 65.6           & 41.4          & 82.9          & 51.9          & 99.1           \\
I2LMeshNet  ~\cite{Moon_I2L_MeshNet}     & ECCV 2020                 & 140.5         & 36.6         & 55.7           & 41.1          & 93.2          & 57.7          & -              \\
TCMR  $\dagger$    ~\cite{TCMR_Choi_2021}         & CVPR 2021                 & -             & -            & 62.3           & 41.1          & 95.0          & 55.8          & 111.5          \\
HybrIK*     ~\cite{hybrik}       & CVPR 2021                 & 27.6        & 12.7         & 57.3           & 36.2          & 75.3          & \textbf{45.2}          & 87.9           \\
ProHMR  ~\cite{prohmr}         & ICCV 2021                 & -             & -            & -              & 41.2          & -             & 59.8          & -              \\
PyMAF    ~\cite{pymaf2021}        & ICCV 2021                 & 45.2          & 10.6         & 57.7           & 40.5          & 92.8          & 58.9          & 110.1          \\
METRO     ~\cite{lin2021metro}       & CVPR 2021                 & 229.2         & 56.6         & 54.0           & 36.7          & 77.1          & 47.9          & 88.2           \\
MeshGraphormer  ~\cite{lin2021_mesh_graphormer}  & ICCV 2021                 & 226.5         & 56.6         & 51.2           & 34.5          & 74.7          & 45.6           & 87.7           \\
DSR     ~\cite{dsr2021}         & ICCV 2021                 & -             & -            & 60.9           & 40.3          & 85.7          & 51.7          & 99.5           \\
TCFormer  ~\cite{TCFormer}  & CVPR 2022                 & -         & -         & 62.9           & 42.8          & 80.6          & 49.3           & -           \\
FastMETRO  ~\cite{cho2022FastMETRO}  & ECCV 2022                 & 48.5         & 15.8         & 53.9           & 37.3          & 77.9          & 48.3           & 90.6           \\
\rowcolor{lightgray}
FeatER           & \multicolumn{1}{c|}{}     & 11.4 & 8.8 & \textbf{49.9}  & \textbf{32.8} &         \textbf{73.4}      &       45.9        &    \textbf{86.9}            \\ \hline
\end{tabular}
}
\label{tab: 3dmesh}
\vspace{-10pt}
\end{table*}

\subsection{Ablation Study}
\label{sec:Ablation}
We conduct the ablation study on COCO, Human3.6M, and 3DPW datasets. The train/test split, experiments setting, and evaluation metrics are the same as in Sections \ref{sec:main_2DHPE} and \ref{sec:main_3DMesh}. 

\textbf{Effectiveness of FeatER:} We compare the vanilla transformer architecture in Fig. \ref{fig:encoder} (a) with our FeatER block in Fig. \ref{fig:encoder} (b) on 2D HPE, 3D HPE, and HMR tasks in Table \ref{tab: ab_2dhpe} and \ref{tab: ab_3dmesh}, respectively. 

`No transformer' means we do not employ transformer to refine the coarse feature maps extracted by the CNN backbone. 
The `VanillaTransformer' indicates that we utilize the vanilla transformer as described in Section \ref{sec:vanillatrans} instead of the proposed FeatER blocks to refine the coarse feature maps in the pipeline. For fair comparisons, given the input of the blocks $X_{in}^{FM} \in \mathbb{R} ^{n \times h \times w} $, FeatER blocks return the output $X_{out}^{FM} \in \mathbb{R} ^{n \times h \times w} $. `VanillaTransformer' first flattens the input to $X_{in} \in \mathbb{R} ^{n \times d} $ and returns $X_{out} \in \mathbb{R} ^{n \times d} $. Next, the flattened output is reshaped to $X_{out}^{FM} \in \mathbb{R} ^{n \times h \times w} $ following the feature map format.
`VanillaTransformer\_S' is the small version of vanilla transformer, which has similar computational complexity with FeatER blocks and the embedding dimension shrinks to $d = 384$. `VanillaTransformer\_L' is the large version of vanilla transformer, which requires more memory and computational costs.  

In Table \ref{tab: ab_2dhpe}, without employing transformer, the network requires fewer Params and MACs but the performance is worse than others. Once transformer is applied, FeatER outperforms VanillaTransformer\_S by a large margin with similar MACs and 42\% of Params. Even compared with VanillaTransformer\_L, FeatER can achieve competitive results while only requiring 12\% Params and 55\% MACs. 


\begin{table}[htp]
\vspace{-5pt}
\centering
  \caption{Ablation study on transformer design for 2D HPE task on COCO validation set. }
  \resizebox{1\linewidth}{!}
  {
  \begin{tabular}{c|ccc|cccccc}
\hline
Model           & Input size & Params (M) & MACs (G) & AP $\uparrow$   & AP50 $\uparrow$ & AP75 $\uparrow$ & AP(M) $\uparrow$ & AP(L) $\uparrow$ & AR $\uparrow$   \\ \hline
No transformer & 256$\times$192     & 7.2       & 4.4      & 72.9 & 87.8 & 79.1 & 69.0  & 78.3  & 76.4 \\
VanillaTransformer\_S & 256$\times$192     & 19.5       & 5.4      & 74.0 & 89.2 & 80.4 & 70.4  & 79.5  & 78.4 \\
VanillaTransformer\_L & 256$\times$192     & 69.1       & 9.8      & 75.1 & 90.2 & 81.3 & 71.0  & 81.8  & 80.0 \\ \hline
FeatER            & 256$\times$192     & 8.1        & 5.4      & 74.9 & 89.8 & 81.6 & 71.2  & 81.7  & 80.0 \\ \hline
\end{tabular}
}
\label{tab: ab_2dhpe}
\vspace{-5pt}
\end{table}

\begin{table}[htp]
\vspace{-10pt}
\centering
  \caption{\small{Ablation study on transformer design for 3D HPE and HMR tasks on Human3.6M and 3DPW datasets.} }
  \resizebox{1\linewidth}{!}
  {
  \begin{tabular}{c|cc|cc|ccc}
\hline
                      &           &         & \multicolumn{2}{c|}{Human3.6M} & \multicolumn{3}{c}{3DPW} \\ \hline
Model                 & Params (M) & MACs (G) & MPJPE $\downarrow$        & PA-MPJPE $\downarrow$        & MPJPE $\downarrow$  & PA-MPJPE $\downarrow$  & MPVE $\downarrow$  \\ \hline
No transformer & 9.7      & 6.9     & 54.4         & 39.1            & 92.5  & 56.9     & 109.6 \\
VanillaTransformer\_S & 30.5      & 8.9     & 52.7         & 37.6            & 90.6  & 55.3     & 107.4 \\
VanillaTransformer\_L & 127.7     & 18.2    &  48.3            &  33.7               &  88.3     &  54.8        &   105.6    \\ \hline
FeatER                  & 11.4      & 8.8     & 49.9         & 32.8            & 88.4  & 54.5     & 105.6 \\ \hline
\end{tabular}
}
\label{tab: ab_3dmesh}
\vspace{-5pt}
\end{table}


In Table \ref{tab: ab_3dmesh}, we observe a similar trend where FeatER surpasses VanillaTransformer\_S by a large margin with similar MACs and 37\% of Params. While only requiring 9\% Params and 48\% MACs, FeatER achieves comparable results compared with using VanillaTransformer\_L.

We can conclude that FeatER is an extremely efficient network with a strong modeling capability, which is more suitable for 2D HPE, 3D HPE, and HMR tasks. More analysis and feature maps visualization
are shown in \textcolor{blue}{Supplementary \ref{sec:supp_Feater}}.


\textbf{Effectiveness of using feature map reconstruction module:}
The purpose of applying the feature map reconstruction module is to improve the generalization ability of our network. In our current design, the Reconstruction Module is for 3D pose and mesh tasks. The occlusion makes 3D HPE and HMR more challenging than 2D HPE. Thus 3D HPE and HMR can be more benefited by adding the Reconstruction Module. As shown in Table \ref{tab: ab_HRM}, once the Reconstruction Module is added, the performance can be improved. 
If we move the Reconstruction Module for 2D HPE, although the performance of 2D HPE can be increased slightly, the performance of 3D HPE and HMR can not be boosted significantly.
If we use two Reconstruction Modules, one for 2D HPE and another for 3D HPE and HMR. The performance also can not be further improved. Moreover, the Params and MACs are increased, which is not what we want. Thus, putting the Feature Map Reconstruction Module in the 3D part is the optimal solution to trade off accuracy and efficiency.
More analysis and results about the feature map reconstruction module are provided in \textcolor{blue}{Supplementary \ref{sec:supp_reconstruction}}.
\begin{table}[htp]
\vspace{-5pt}
\centering
  \caption{Ablation study on the different positions of the Feature Map Reconstruction Module. }
  \resizebox{1\linewidth}{!}
  {
\begin{tabular}{ccc|cc|c|c}
\hline
                                                               & \multicolumn{1}{l}{}           & \multicolumn{1}{l|}{}         & \multicolumn{2}{c|}{COCO} & \textbf{Human3.6M} & \textbf{3DPW} \\ \hline
\multicolumn{1}{c|}{}                                          & \multicolumn{1}{l|}{Params (M)} & \multicolumn{1}{l|}{MACs (G)} & AP $\uparrow$         & AR $\uparrow$           & MPJPE $\downarrow$       & MPVE $\downarrow$   \\ \hline
\multicolumn{1}{c|}{No Reconstruction Module}                            & \multicolumn{1}{c|}{10.4}      & 7.7                           & 74.9        & 80.0        & 53.3      & 94.5 \\ \hline
\multicolumn{1}{c|}{In the 2D Part}                            & \multicolumn{1}{c|}{11.4}      & 8.8                           & 75.3        & 80.2        & 52.8      & 88.7 \\ \hline
\multicolumn{1}{c|}{In the 3D Part (FeatER's design)}                            & \multicolumn{1}{c|}{11.4}      & 8.8                           & 74.9        & 80.0        & 49.9      & 86.9 \\ \hline
\multicolumn{1}{c|}{In both the 2D part and 3D part} & \multicolumn{1}{c|}{12.5}      & 10.0                          & 75.3        & 80.2        & 49.8      & 87.1 \\ \hline
\end{tabular}
}
\label{tab: ab_HRM}
\vspace{-10pt}
\end{table}


\subsection{Qualitative Results}
\begin{figure*}[htp]
\vspace{-5pt}
  \centering
  \includegraphics[width=0.99\linewidth]{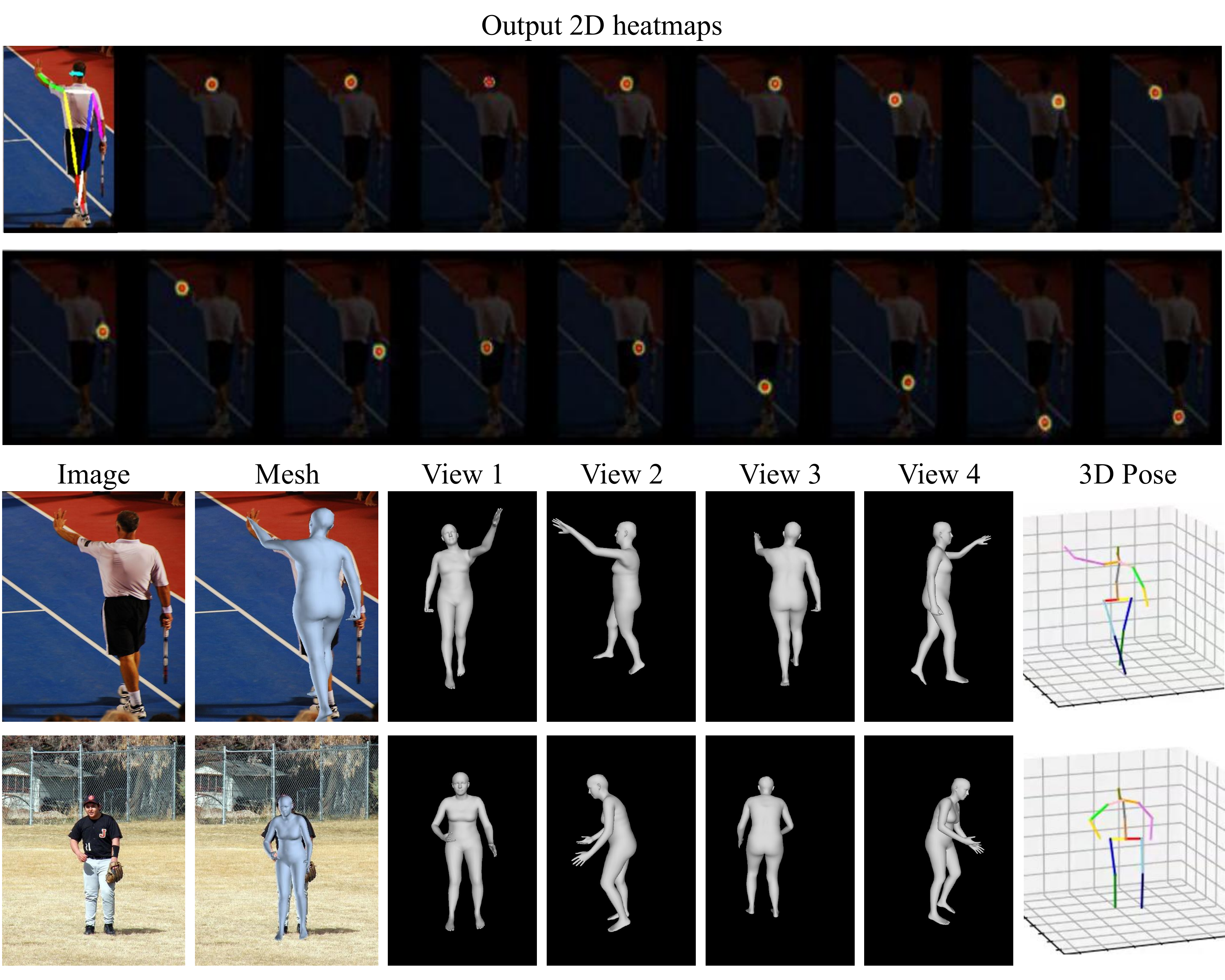}
  \vspace{-10pt}
  \caption{Qualitative results of the proposed FeatER. Images are taken from the in-the-wild COCO \cite{lin2014mscoco} dataset.}
  \label{fig:vis1}
  \vspace{-10pt}
\end{figure*}

\begin{figure}[h!]
\vspace{-5pt}
  \centering
  \includegraphics[width=0.99\linewidth]{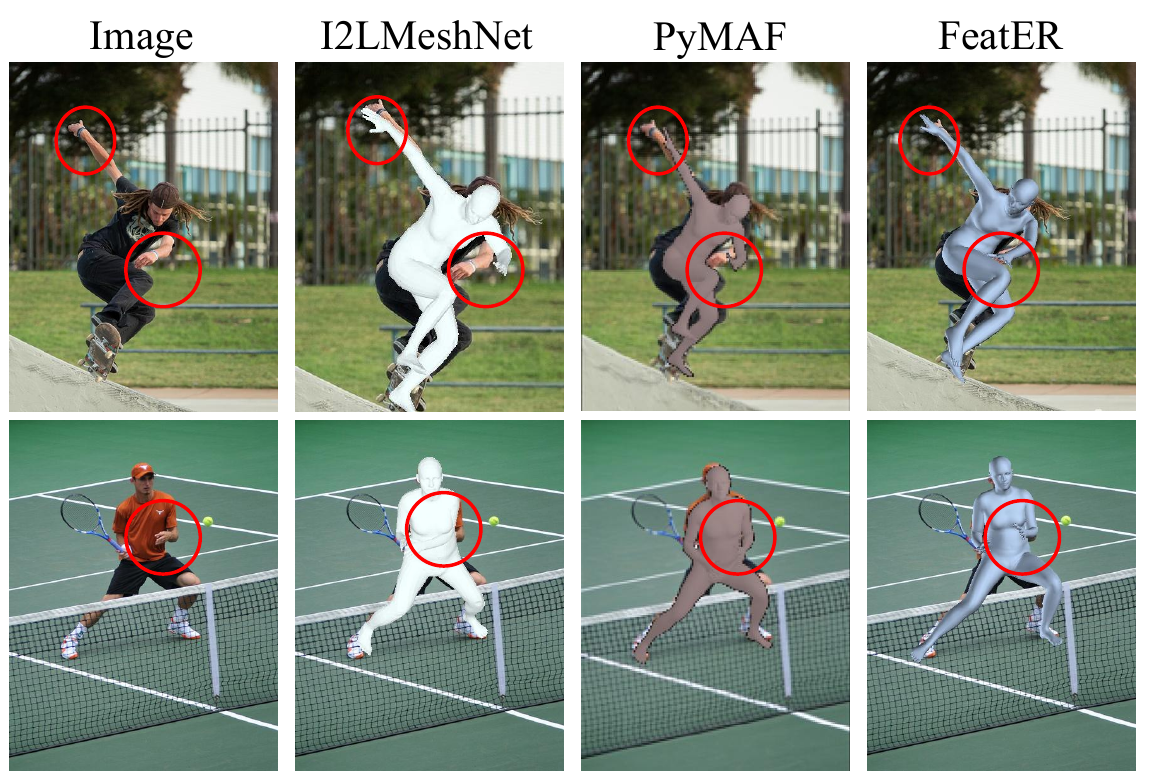}
  \vspace{-5pt}
  \caption{Qualitative comparison with other methods. Images are taken from the in-the-wild COCO \cite{lin2014mscoco} dataset. The red circles highlight locations where FeatER is more accurate than others. We follow previous work \cite{lin2021metro,lin2021_mesh_graphormer,Moon_I2L_MeshNet,pymaf2021} to visualize human mesh using the SMPL \textit{gender neutral} model.}
  \label{fig:vis2}
  \vspace{-5pt}
\end{figure}

To show the qualitative results of the proposed FeatER for human reconstruction (2D HPE, 3D HPE, and HMR), we use the challenging COCO dataset which consists of in-the-wild images. Given various input images, FeatER can estimate reliable human poses and meshes as shown in Fig.~\ref{fig:vis1}. When comparing with I2LMeshNet \cite{Moon_I2L_MeshNet} and PyMAF \cite{pymaf2021} in Fig. \ref{fig:vis2}, the areas highlighted by red circles indicate that FeatER outputs more accurate meshes under challenging scenarios. We provide more visual examples on different datasets in \textcolor{blue}{Supplementary \ref{sec:supp_Qualitative}}. 

\section{Conclusion}
\label{sec:Conclusions}

In this paper, we present FeatER, a novel feature map-based transformer architecture for HPE and HMR. FeatER can preserve the feature map representations and effectively model global correlations between them via self-attention. By performing decomposition with th $w$ and $h$ dimensions, FeatER significantly reduces the computational complexity compared with vanilla transformer architecture. Furthermore, the introduced feature map reconstruction module improves the robustness of the estimated human pose and mesh. Extensive experiments show that FeatER improves performance while significantly reducing the computational cost for HPE and HMR tasks. 

While our  network does not raise a direct negative societal impact, it may be used by some applications for malicious purposes, such as unwarranted surveillance. To avoid possible negative societal impact, we urge the readers to limit our network to ethical and legal use-cases.

\textbf{Acknowledgement}: This work is supported by the National Science Foundation under Grant No. 1910844.



\newpage

\newpage
\noindent \textbf{\large{Supplementary Material}}

The supplementary material is organized into the following sections:

\begin{itemize}

\item Section \ref{sec:supp_Complexity}: Details of Complexity Comparison.

\item Section \ref{sec:supp_Feater}: Effectiveness of FeatER by Feature Maps Visualization.

\item Section \ref{sec:supp_reconstruction}: Effectiveness of Using the Feature Map Reconstruction Module.

\item Section \ref{sec:supp_Lifting}: 2D-3D Lifting Module.

\item Section \ref{sec:supp_loss}: Loss Function.

\item Section \ref{sec:supp_Qualitative}: More Qualitative Results.

\item Section \ref{Broader}: Broader Impact and Limitations.

\end{itemize}

\appendix




\section{Details of Complexity Comparison}
\label{sec:supp_Complexity}


In Table \ref{tab: appd_complexity}, we list the layer-by-layer comparison between one vanilla transformer block and one FeatER block. The shape of a stack of feature maps is $[n,h,w]$, where $n$ is the number of feature maps, $h$ and $w$ is the height and width of the feature maps, respectively. If $h=w=64$, the embedding dimension of $d$ would be $d=hw=4096$ without discarding any information. Since $d$ is much larger than $n$, the computational complexity of one vanilla transformer block and one FeatER block can be written as $\mathcal{O}(d^2)$ and $\mathcal{O}(d^{3/2})$, respectively. 

To be more specific, let there be a stack of 32 feature maps with the dimension of $[32,64,64]$. One vanilla transformer block requires 4.3G MACs when the embedding dimension is $d=64\times64=4096$ (i.e., flattening the spatial dimension). 
Even if we further reduce the embedding dimension to $d=1024$, it still needs 0.27G MACs. 
However, given feature maps $[32,64,64]$, FeatER only requires 0.09G MACs, which significantly reduces the computational cost. 

\begin{table*}[htp]
\scriptsize
\renewcommand\arraystretch{1.4}
\centering
  \caption{The detailed complexity comparison between one vanilla transformer block and one FeatER block. We calculate their MACs based on the input and output with the corresponding operation. }
  \resizebox{1\linewidth}{!}
  {
\begin{tabular}{|ccccc|ccccc|}
\hline
\rowcolor{lightgray}\multicolumn{5}{|c|}{\textbf{Vanilla Transformer block}}                                                                                                                                                            & \multicolumn{5}{c|}{\textbf{FeatER block}}                                                                                                                                                                \\ \hline
\multicolumn{5}{|c|}{\textbf{Attention Layer:}}                                                                                                                                                                     & \multicolumn{5}{c|}{\textbf{Attention Layer(AttentionW):}}                                                                                                                                                    \\ \hline
\multicolumn{1}{|c|}{description}  & \multicolumn{1}{c|}{input}                      & \multicolumn{1}{c|}{output}            & \multicolumn{1}{c|}{operation}        & MACs                                        & \multicolumn{1}{c|}{description}  & \multicolumn{1}{c|}{input}                          & \multicolumn{1}{c|}{output}              & \multicolumn{1}{c|}{operation}          & MACs                       \\ \hline
\multicolumn{1}{|c|}{x to QKV}    & \multicolumn{1}{c|}{$x_{in}$ $[n,d]$}               & \multicolumn{1}{c|}{$QKV$ $[n,3d]$}    & \multicolumn{1}{c|}{nn.Linear(d, 3d)} & \textcolor{orange}{$3nd^2$}                     & \multicolumn{1}{c|}{x to QKV}     & \multicolumn{1}{c|}{$x_{in}^{w}$: $[n,h,w]$}                 & \multicolumn{1}{c|}{$QKV$ $[n,h,3w]$}    & \multicolumn{1}{c|}{nn.Linear(w, 3w)}   & \textcolor{orange}{$3nhw^2$} \\ \hline
\multicolumn{1}{|c|}{$a_1 = QK^T$} & \multicolumn{1}{c|}{$Q[n,d]$, $K^T[d,n]$} & \multicolumn{1}{c|}{$a_1 [n,n]$}      & \multicolumn{1}{c|}{torch.matmul}     & \textcolor{orange}{$n^{2}d$ }            & \multicolumn{1}{c|}{$a_1^w = Q^w {K^w}^T$} & \multicolumn{1}{c|}{$Q^w[h,n,w]$,${K^w}^T[h,w,n]$} & \multicolumn{1}{c|}{$a_1^w$ $[h,n,n]$}      & \multicolumn{1}{c|}{torch.matmul}       & \textcolor{orange}{$n^2hw$}   \\ \hline
\multicolumn{1}{|c|}{$x_{attn} = a_1 V$}  & \multicolumn{1}{c|}{$a_1[n,n]$, $V[n,d]$ } & \multicolumn{1}{c|}{$x_{attn} [n,d]$} & \multicolumn{1}{c|}{torch.matmul}     & \textcolor{orange}{$n^{2}d$}           & \multicolumn{1}{c|}{$x_{attn}^{w} = a_1^w V^w $}  & \multicolumn{1}{c|}{$a_1^w [h,n,n]$, $V [h,n,w]$} & \multicolumn{1}{c|}{$x_{attn}^w [h,n,w]$} & \multicolumn{1}{c|}{torch.matmul}       & \textcolor{orange}{$n^2hw$}  \\ \hline
\multicolumn{1}{|l|}{}             & \multicolumn{1}{c|}{}                           & \multicolumn{1}{c|}{}                  & \multicolumn{1}{c|}{}                 &                                             & \multicolumn{5}{c|}{\textbf{Attention Layer(AttentionH):}}                                                                                                                                                    \\ \cline{6-10} 
\multicolumn{1}{|l|}{}             & \multicolumn{1}{l|}{}                           & \multicolumn{1}{l|}{}                  & \multicolumn{1}{l|}{}                 &                                             & \multicolumn{1}{c|}{description}  & \multicolumn{1}{c|}{input}                          & \multicolumn{1}{c|}{output}              & \multicolumn{1}{c|}{operation}          & MACs                       \\ \cline{6-10} 
\multicolumn{1}{|l|}{}             & \multicolumn{1}{c|}{}                           & \multicolumn{1}{c|}{}                  & \multicolumn{1}{c|}{}                 &                                             & \multicolumn{1}{c|}{x to QKV}     & \multicolumn{1}{c|}{$x_{in}^{h}$: $[n,w,h]$}                 & \multicolumn{1}{c|}{$QKV$ $[n,w,3h]$}    & \multicolumn{1}{c|}{nn.Linear(h, 3h)}   & \textcolor{orange}{$3nh^{2}w$} \\ \cline{6-10} 
\multicolumn{1}{|l|}{}             & \multicolumn{1}{c|}{}                           & \multicolumn{1}{c|}{}                  & \multicolumn{1}{c|}{}                 &                                             & \multicolumn{1}{c|}{$a_1^h = Q^h {K^h}^T$} & \multicolumn{1}{c|}{$Q^h[w,n,h]$,${K^h}^T[w,h,n]$} & \multicolumn{1}{c|}{$a_1^h$ $[w,n,n]$}      & \multicolumn{1}{c|}{torch.matmul}       & \textcolor{orange}{$n^2hw$}   \\ \cline{6-10} 
\multicolumn{1}{|c|}{}             & \multicolumn{1}{c|}{}                           & \multicolumn{1}{c|}{}                  & \multicolumn{1}{c|}{}                 &                                             & \multicolumn{1}{c|}{$x_{attn}^{h} = a_1^h V^h $}  & \multicolumn{1}{c|}{$a_1^h$ $[w,n,n]$,$V^h$ $[w,n,h]$} & \multicolumn{1}{c|}{$x_{attn}^w [w,n,h]$} & \multicolumn{1}{c|}{torch.matmul}       & \textcolor{orange}{$n^2hw$}   \\ \hline
\multicolumn{5}{|c|}{\textbf{Projection Layer}}                                                                                                                                                                     & \multicolumn{5}{c|}{\textbf{Projection Layer}}                                                                                                                                                            \\ \hline
\multicolumn{1}{|c|}{description}  & \multicolumn{1}{c|}{input}                      & \multicolumn{1}{c|}{output}            & \multicolumn{1}{c|}{operation}        & MACs                                        & \multicolumn{1}{c|}{description}  & \multicolumn{1}{c|}{input}                          & \multicolumn{1}{c|}{output}              & \multicolumn{1}{c|}{operation}          & MACs                       \\ \hline
\multicolumn{1}{|c|}{projection:}  & \multicolumn{1}{c|}{$x_{attn} [n,d]$}          & \multicolumn{1}{c|}{$x [n,d]$}       & \multicolumn{1}{c|}{nn.Linear(d, d)}  & \textcolor{orange}{$nd^2$}                  & \multicolumn{1}{c|}{projection:}  & \multicolumn{1}{c|}{  $x_{attn}^{FM} [n,h,w]$}            & \multicolumn{1}{c|}{ $x [n,h,w]$}       & \multicolumn{1}{c|}{nn.Conv2d(n, n,1)}  &  \textcolor{orange}{$n^2hw$}    \\ \hline
\multicolumn{5}{|c|}{\textbf{MLP Layer (mlp raito=2) in FFN:}}                                                                                                                                                             & \multicolumn{5}{c|}{\textbf{CONV Layer (conv raito=2) in FFN:}}                                                                                                                                                  \\ \hline
\multicolumn{1}{|c|}{description}  & \multicolumn{1}{c|}{input}                      & \multicolumn{1}{c|}{output}            & \multicolumn{1}{c|}{operation}        & MACs                                        & \multicolumn{1}{c|}{description}  & \multicolumn{1}{c|}{input}                          & \multicolumn{1}{c|}{output}              & \multicolumn{1}{c|}{operation}          & MACs                       \\ \hline
\multicolumn{1}{|c|}{MLP}          & \multicolumn{1}{c|}{$x [n,d]$}                & \multicolumn{1}{c|}{$x_{hidden} [n,2d]$}  & \multicolumn{1}{l|}{nn.Linear(d,2d)}  & \textcolor{orange}{$2nd^2$} & \multicolumn{1}{c|}{CONV}          & \multicolumn{1}{c|}{ $x [n,h,w]$ }                  & \multicolumn{1}{c|}{$x_{hidden} [2n,h,w]$}  & \multicolumn{1}{l|}{nn.Conv2d(n, 2n,1)} & \textcolor{orange}{$2n^{2}hw$} \\ \hline
\multicolumn{1}{|c|}{MLP}          & \multicolumn{1}{c|}{$x_{hidden} [n,2d]$}           & \multicolumn{1}{c|}{$x [n,d]$}       & \multicolumn{1}{l|}{nn.Linear(2d,d)}  & \textcolor{orange}{$2nd^2$}  & \multicolumn{1}{c|}{CONV}          & \multicolumn{1}{c|}{$x_{hidden} [2n,h,w]$}             & \multicolumn{1}{c|}{$x [n,h,w]$}       & \multicolumn{1}{l|}{nn.Conv2d(2n, n,1)} & \textcolor{orange}{$2n^{2}hw$} \\ \hline
\rowcolor{lightgray}\multicolumn{5}{|c|}{\textbf{Total: \textcolor{red}{$8nd^2 + 2n^{2}d$} }}                                                                                                                              & \multicolumn{5}{c|}{\textbf{Total: \textcolor{red}{$3nhw(h+w) + 9n^2(wh)$}}}                                                                                                                               \\ \hline
\rowcolor{lightgray}
\textbf{}                          & \textbf{}                                       & \textbf{}                              & \textbf{}                             & \textbf{}                                   & \multicolumn{5}{c|}{\textbf{Total: \textcolor{red}{$6nd^{3/2} + 9n^{2}d$ }} when $w*h=d$ and $w=h$}                                                                                                                  \\ \hline
\end{tabular}
}
\label{tab: appd_complexity}
\end{table*}

\section{Effectiveness of FeatER by Feature Maps Visualization}
\label{sec:supp_Feater}

\begin{figure*}[htp]
  \centering
  \includegraphics[width=0.8\linewidth]{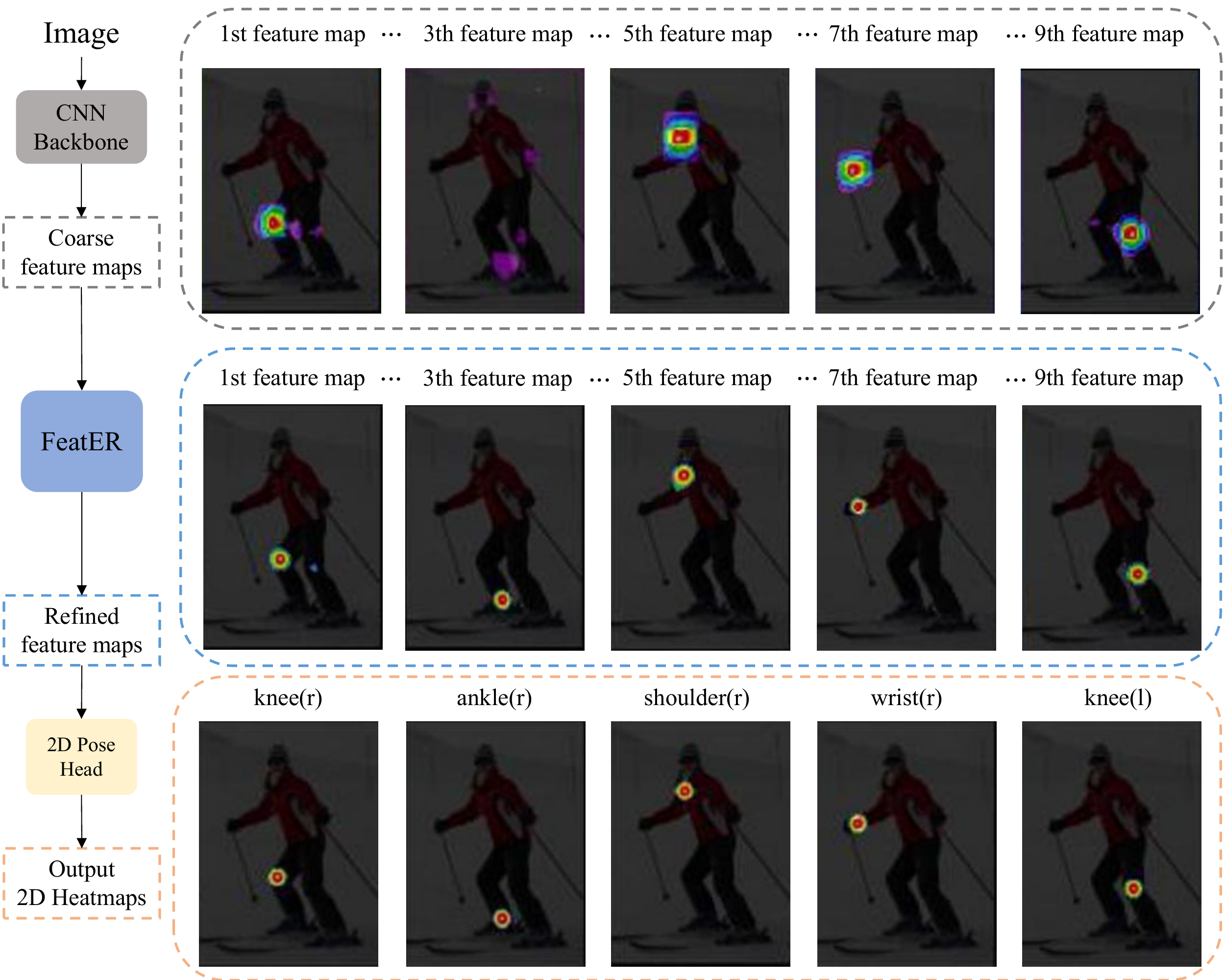}
  \caption{Visualization of coarse feature maps (extracted by CNN backbone) and refined feature maps (refined by FeatER). }
  \vspace{-5pt}
  \label{fig: feature_map}
\end{figure*}

We visualize the coarse feature maps (extracted by CNN backbone) and the refined feature maps (refined by FeatER) in Fig. \ref{fig: feature_map}. These examples demonstrate that our proposed feature map-based transformer (FeatER) blocks can successfully refine the coarse feature maps by predicting more accurate joint locations, thereby improving the performance of human reconstruction tasks (2D HPE, 3D HPE, and HMR). 

\section{Effectiveness of Using the Feature Map Reconstruction Module}
\label{sec:supp_reconstruction}

We compare the performance of our network with and without the feature map reconstruction module in Table \ref{tab: ab_action}. The performance is improved in all cases, including for the most challenging actions on the Human3.6M indoor dataset with heavy occlusions such as Photo, SitD (sitting down), and WalkD (walking with dog). The feature map reconstruction module effectively reduces the error by 4.4, 3.7, and 4.6 for these actions, respectively. Then, we compare the results on the in-the-wild 3DPW dataset, the MPJPE and MPVE also have decreased. Therefore, through this analysis, we validate the effectiveness of using the feature map reconstruction module.
\begin{table*}[htp]
\centering
  \caption{Ablation study on the effectiveness of using our feature map reconstruction module on Human3.6M. `FM-Rec' means Feature Map Reconstruction Module and `$\Delta$' denotes the performance improvement.  }
  \resizebox{0.8\linewidth}{!}
  {
\begin{tabular}{c|cccccccccc|cc}
\hline
            & \multicolumn{10}{c|}{Human3.6M}                                               & \multicolumn{2}{c}{3DPW} \\ \hline
            & \multicolumn{10}{c|}{MPJPE $\downarrow$ }                                                   & MPJPE $\downarrow$       & MPVE $\downarrow$       \\ \hline
actions     & Dire. & Eat  & Phone & Photo & Pose & Purch. & SitD. & WalkD. & Smoke & Avg. & Avg.        & {Avg.}       \\ \hline
w/o FM-Rec  & 50.1  & 49.5 & 56.8   & 60.0  & 46.3 & 51.0   & 69.4  & 57.8   & 52.4   & 53.3 & 89.9        & 106.9      \\
with FM-Rec & 46.3  & 45.7 & 54.7  & 55.6  & 43.0 & 47.2   & 65.7  & 53.2   & 49.6   & 49.9 & 88.4        & 105.6      \\
\rowcolor{lightgray}
$\Delta$    & 3.8   & 3.8  & 2.1    & 4.4   & 3.3  & 3.8    & 3.7   & 4.6    & 2.8    & 3.4  & 1.5         & 1.3        \\ \hline
\end{tabular}
}

\label{tab: ab_action}
\vspace{-5pt}
\end{table*}

\begin{figure}[htp]
\vspace{-5pt}
  \centering
  \includegraphics[width=0.9\linewidth]{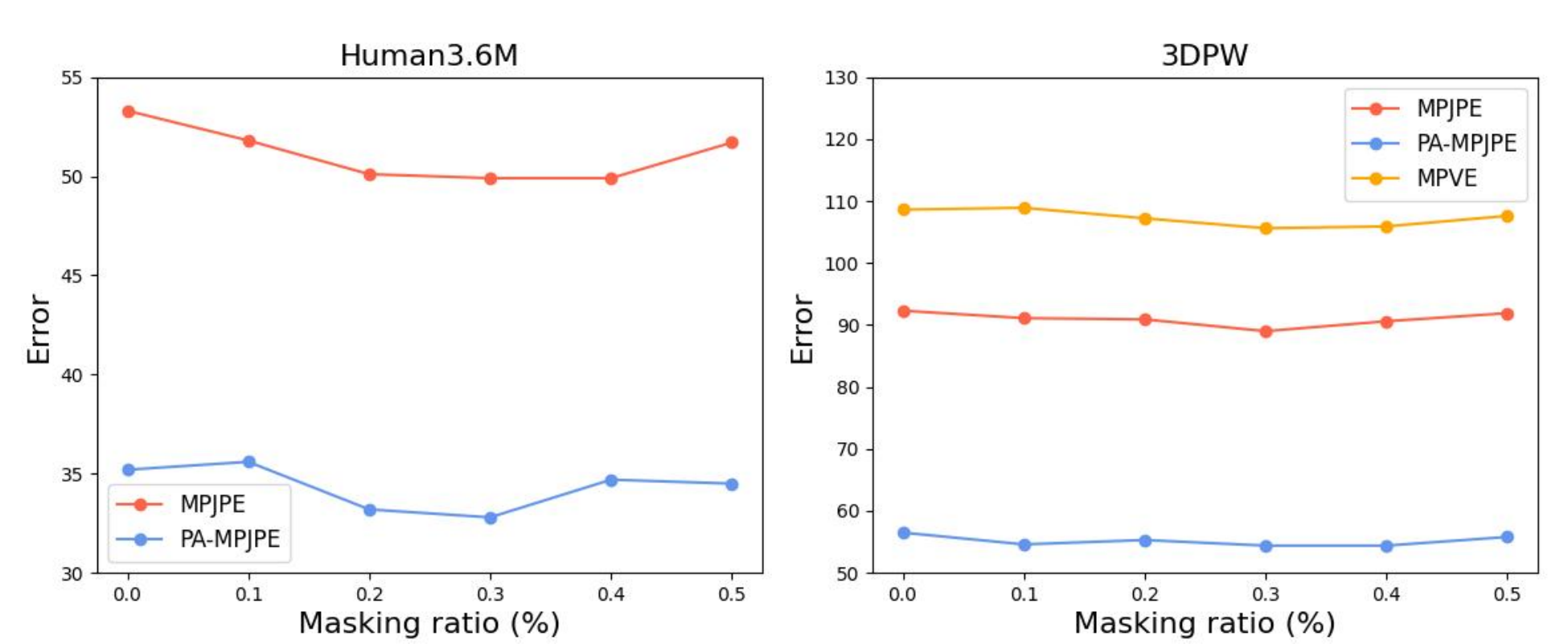}
  \caption{Evaluation of different masking ratios in the feature map reconstruction module. }
  \vspace{-10pt}
  \label{fig:maskrate}
\end{figure}

Next, we investigate the best masking ratio in the feature map reconstruction module. We plot the relations between the error (MPJPE, PA-MPJPE, and MPVE) with the masking ratio in Fig. \ref{fig:maskrate}. We set the masking ratio to be 0.3 since it provides the best results on both Human3.6M and 3DPW datasets. 

\section{2D-3D Lifting Module}
\label{sec:supp_Lifting}

\begin{figure}[htp]
  \centering
  \includegraphics[width=0.8\linewidth]{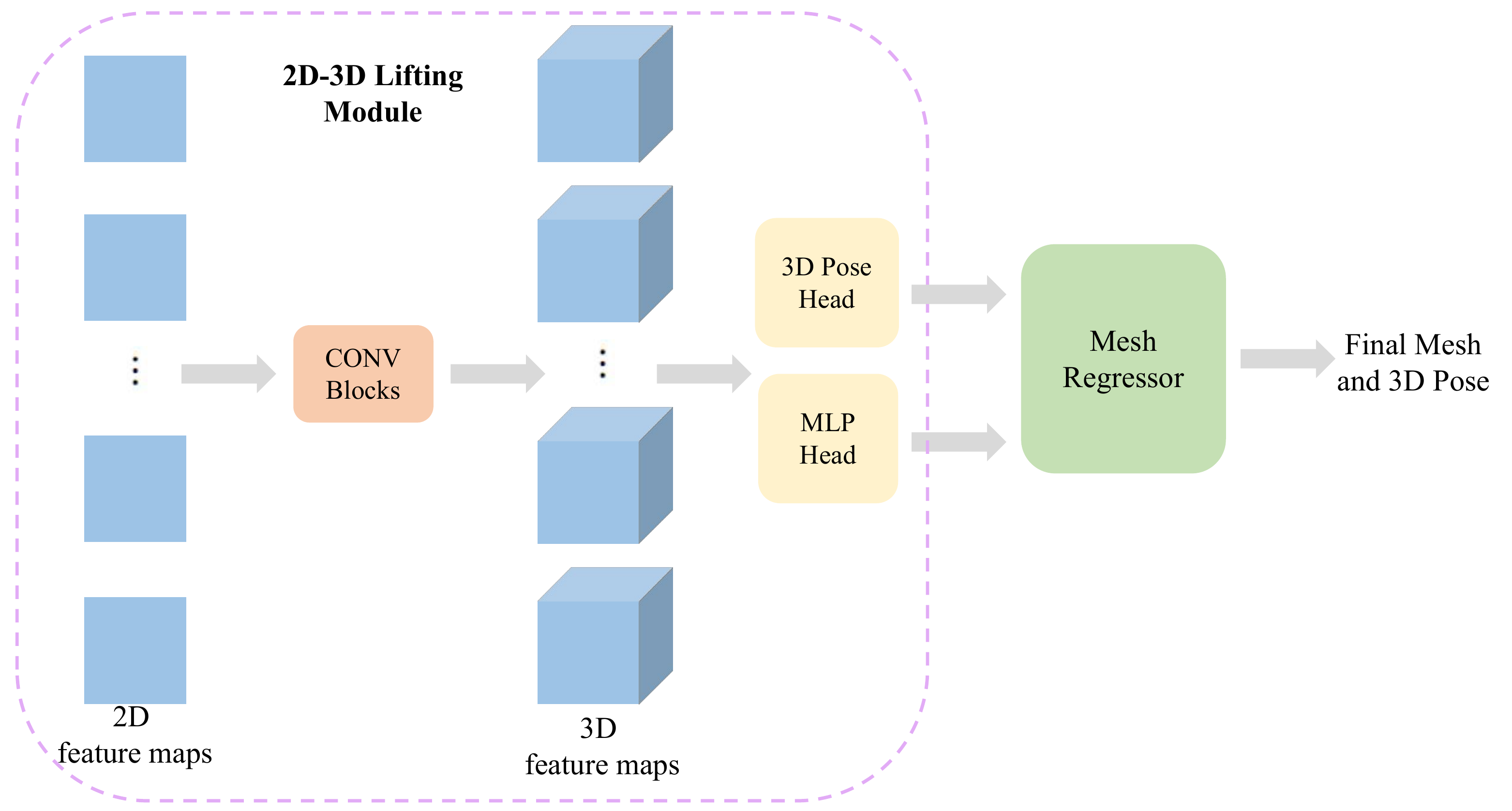}
  \caption{The architecture of the 2D-3D Lifting Module }
  \label{fig:supp_lifting}
\end{figure}

The 2D-3D Lifting module is aimed to lift the 2D feature maps $[n,h,w]$ to 3D feature maps $[n,h,w,d]$. The intermediate 3D Pose can be obtained by a 3D pose head. The MLP head outputs the parameters for the mesh regressor.  The architecture of the 2D-3D Lifting Module is shown in Fig.~\ref{fig:supp_lifting}.


\section{Loss Function}
\label{sec:supp_loss}
\textbf{2D HPE}

We first train our FeatER on COCO dataset for the 2D HPE task. Following \cite{hrnet,tokenpose2021}, we apply the Mean Squared Loss (MSE) between the predicted heatmaps ($HM$)  $HM \in  \mathbb{R} ^{K \times h \times w}$ and the ground truth 2D pose $HM^{GT} \in  \mathbb{R} ^{K \times h \times w}$, where $K$ is the number of joints, $h$ and $w$ are the height and width of heatmaps, respectively. When the input image is $256 \times 192$ and the number of joints is $K=17$, the heatmap size would be $w=64$, and $h=48$, respectively. The MSE for the 2D pose is defined as follows:
\begin{align}
\small
    \mathcal{L}_{2D-Pose} =\| HM - HM^{GT} \|^2
\end{align}

\textbf{3D HPE and HMR}

We apply an $L$1 loss between the predicted 3D pose $J \in  \mathbb{R} ^{K \times 3}$ and the ground truth 3D pose $J_{GT} \in  \mathbb{R} ^{K \times 3}$ following \cite{Choi_2020_ECCV_Pose2Mesh,lin2021metro,hybrik}. $K$ is the number of joints. 
\begin{align}
\small
    \mathcal{L}_{3D-Pose} = \frac{1}{K}\sum_{i=1}^{K}\| J_i - J^{GT}_i \|_1
\end{align}

Following ~\cite{hybrik}, we use the SMPL \cite{SMPL:2015} model to output human mesh, which is obtained by fitting the 3D pose $J$, the shape parameter $\beta$, and the rotation parameter $\theta$ into the SMPL model. We supervise the shape and rotation parameters by applying the $L$2 loss following \cite{SMPL:2015}. The reconstruction loss $\mathcal{L}_{rec} $ is the Mean Square Error (MSE) between the target feature maps and reconstructed  feature maps. The overall loss is defined as follows:

\begin{equation}
\begin{aligned} 
  \begin{aligned}
\mathcal{L}_{overall} &= \mathcal{L}_{3D-Pose} + w_1 \| \beta - \beta^{GT}  \|\\
  &\qquad + w_2 \| \theta - \theta^{GT}  \| + w_3 \mathcal{L}_{rec} 
  \end{aligned}\\
\end{aligned}
\end{equation}

where $w_1=0.01$, $w_2=0.01$ and $w_3=0.005$ are the weights for the loss terms.

\section{More Qualitative Results}
\label{sec:supp_Qualitative}
\begin{figure*}[htp]
  \centering
  \includegraphics[width=0.98\linewidth]{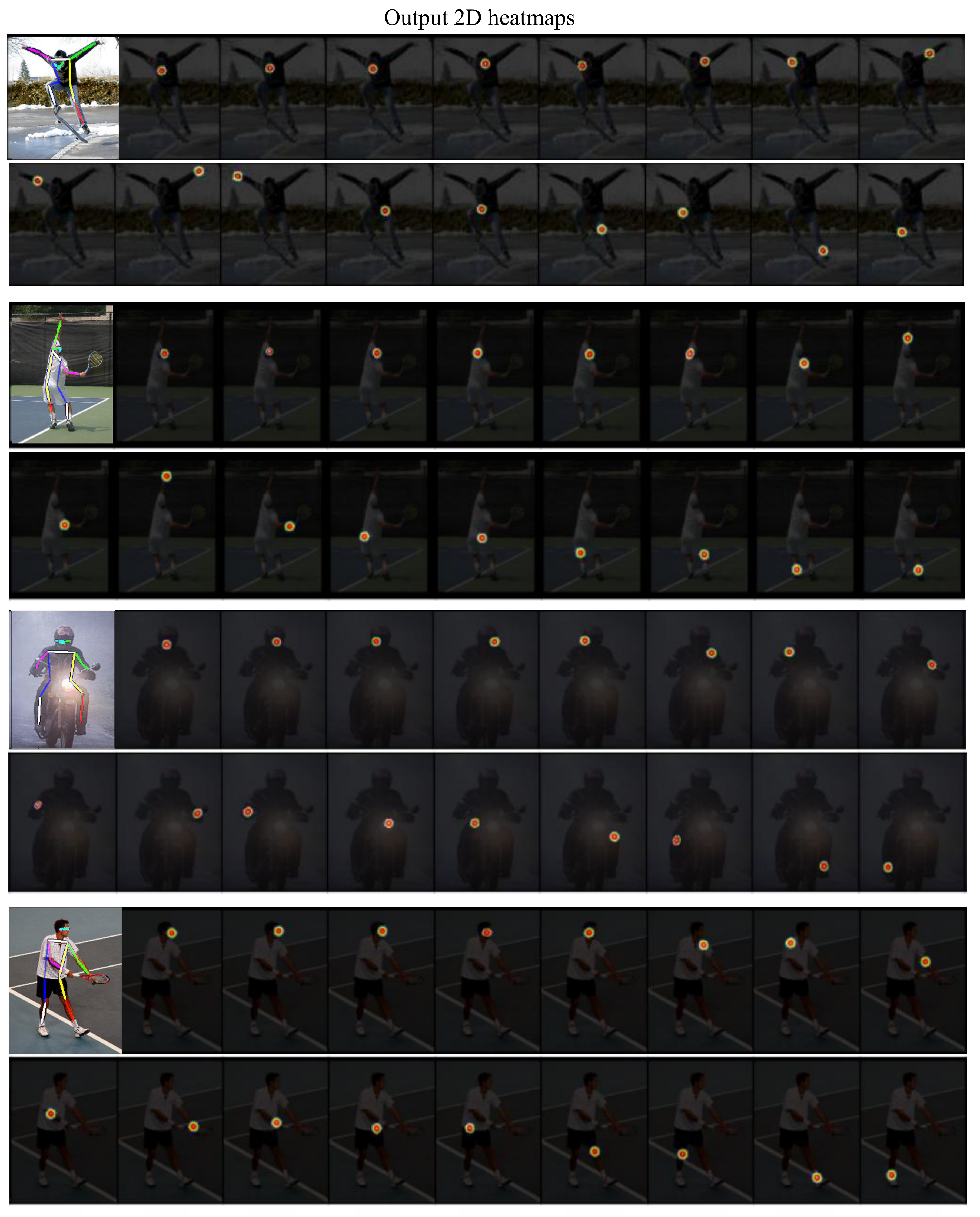}
  \caption{2D heatmaps visualization of the proposed FeatER. Images are taken from the COCO validation set \cite{lin2014mscoco}.}
  \label{fig:supp_2d}
\end{figure*}

\textbf{2D Heatmap and Human Mesh Reconstruction (HMR) Visualization}

\begin{figure*}[htp]
\vspace{-5pt}
  \centering
  \includegraphics[width=0.98\linewidth]{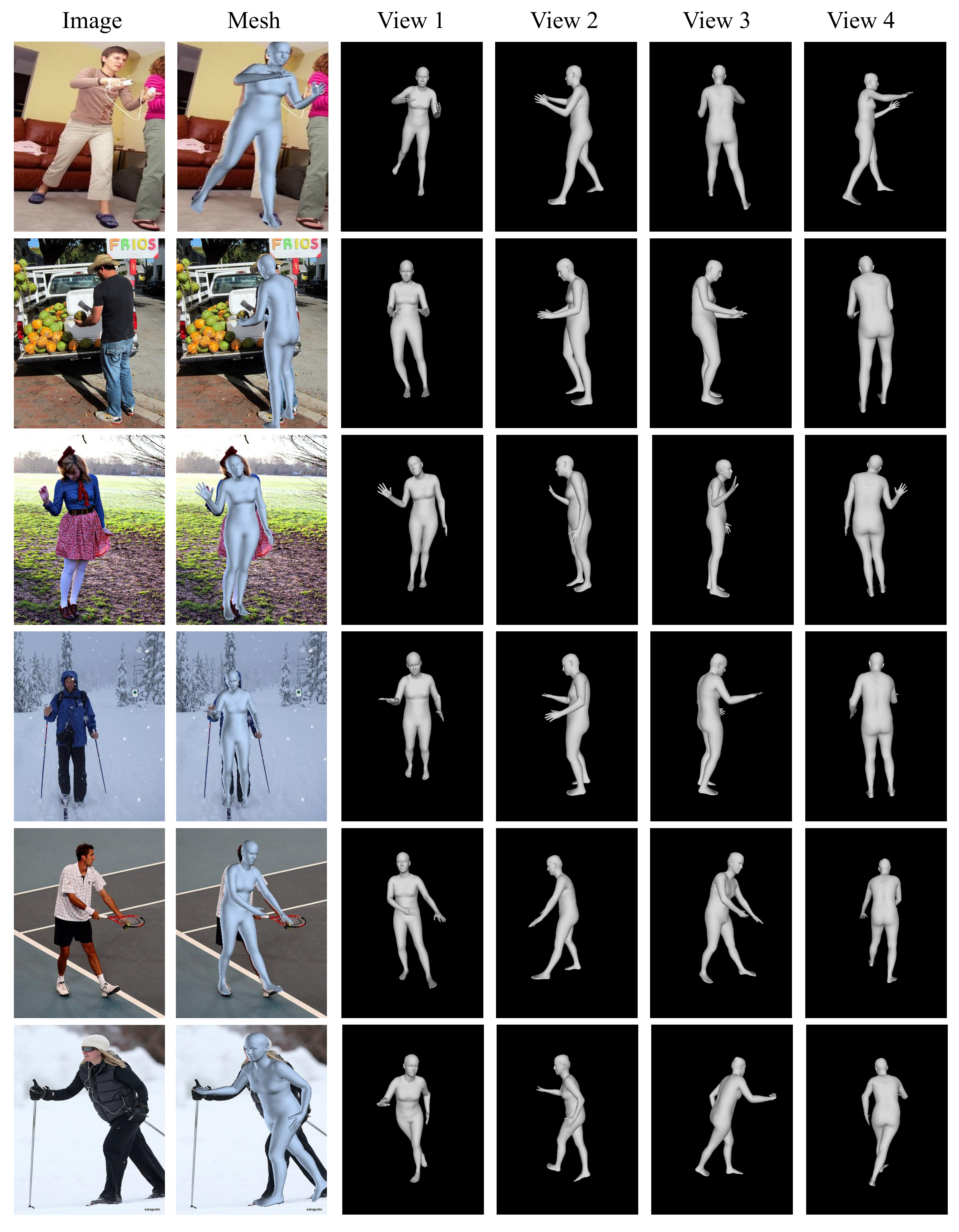}
  \caption{Mesh reconstruction qualitative results of the proposed FeatER. Images are taken from the in-the-wild COCO \cite{lin2014mscoco} dataset.}
  \label{fig:supp_easy}
  \vspace{-5pt}
\end{figure*}

\begin{figure}[htp]
\vspace{-5pt}
  \centering
  \includegraphics[width=0.98\linewidth]{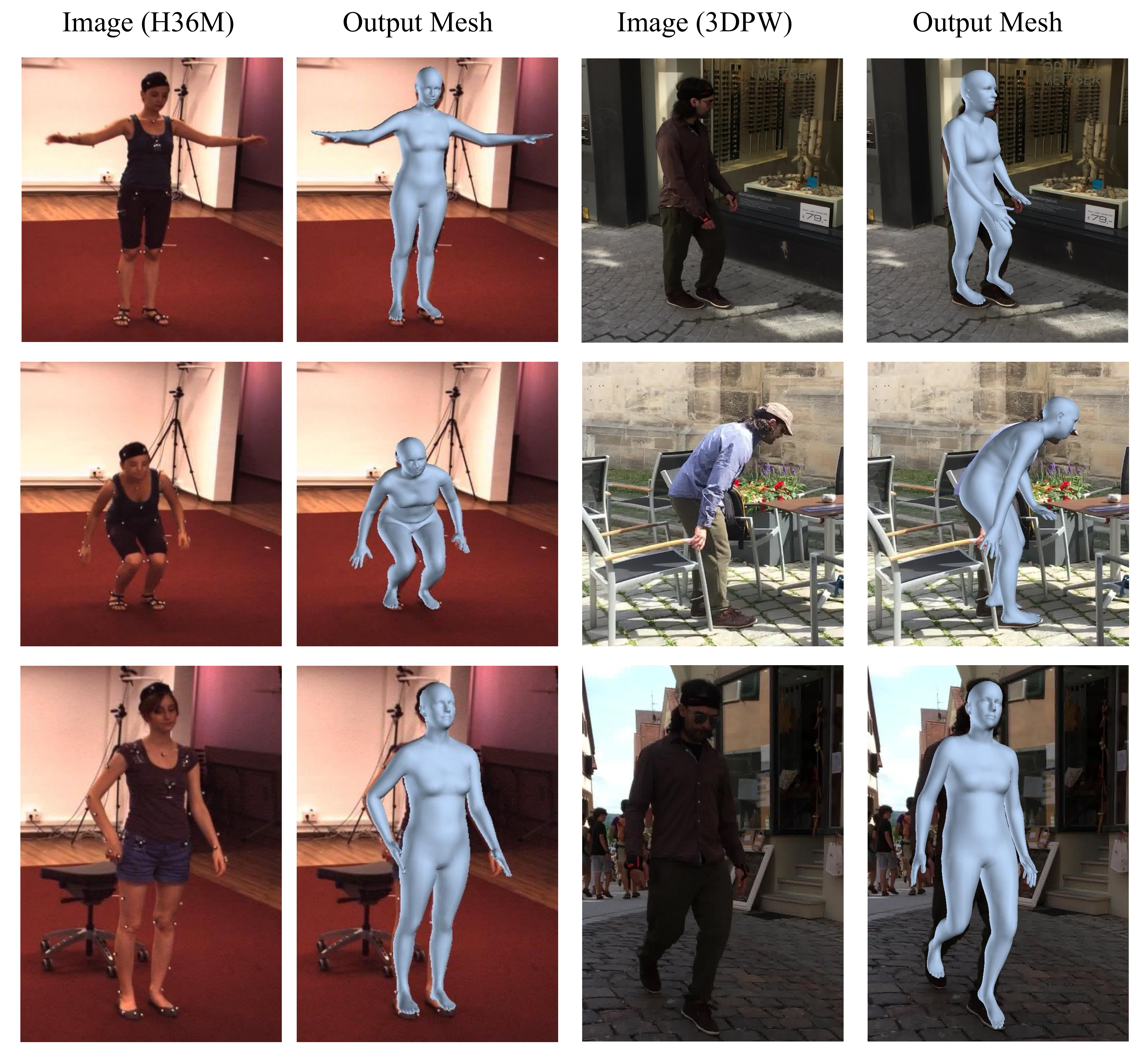}
  \caption{Mesh reconstruction qualitative results of the proposed FeatER. Images are taken from the Human3.6M dataset and 3DPW dataset.}
  \label{fig:supp_3dpw}
  \vspace{-5pt}
\end{figure}

\begin{figure}[htp]
\vspace{-5pt}
  \centering
  \includegraphics[width=0.98\linewidth]{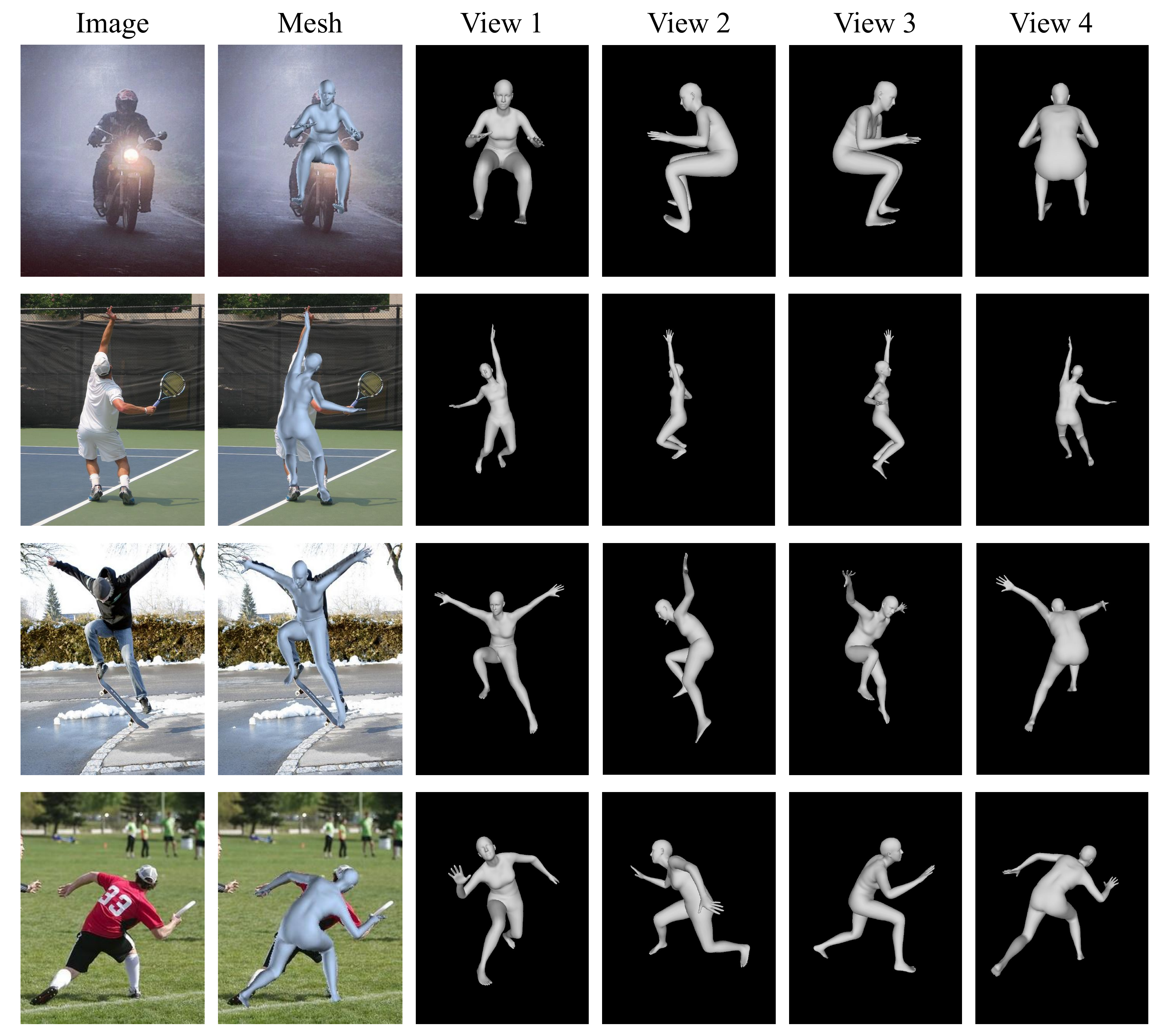}
  \caption{Mesh reconstruction qualitative results of the proposed FeatER for more challenging cases. Images are taken from the in-the-wild COCO \cite{lin2014mscoco} dataset.}
  \label{fig:supp_hard}
  \vspace{-5pt}
\end{figure}

Fig.~\ref{fig:supp_2d} provides visualization of 17 heatmaps (COCO \cite{lin2014mscoco} 17 joints format) and the predicted 2D poses of the input images. The visualization of Human3.6M and 3DPW dataset are shown in Figs.~\ref{fig:supp_3dpw}. Figs.~\ref{fig:supp_easy} and ~\ref{fig:supp_hard} show the HMR visualization of FeatER on several in-the-wild images from the COCO \cite{lin2014mscoco} dataset. FeatER can estimate accurate human meshes of the given images with regular human articulation in Fig.~\ref{fig:supp_easy}. For some very challenging cases, as shown in Fig. \ref{fig:supp_hard}, FeatER can still output reliable human meshes.

\begin{figure}[htp]
\vspace{-5pt}
  \centering
  \includegraphics[width=0.98\linewidth]{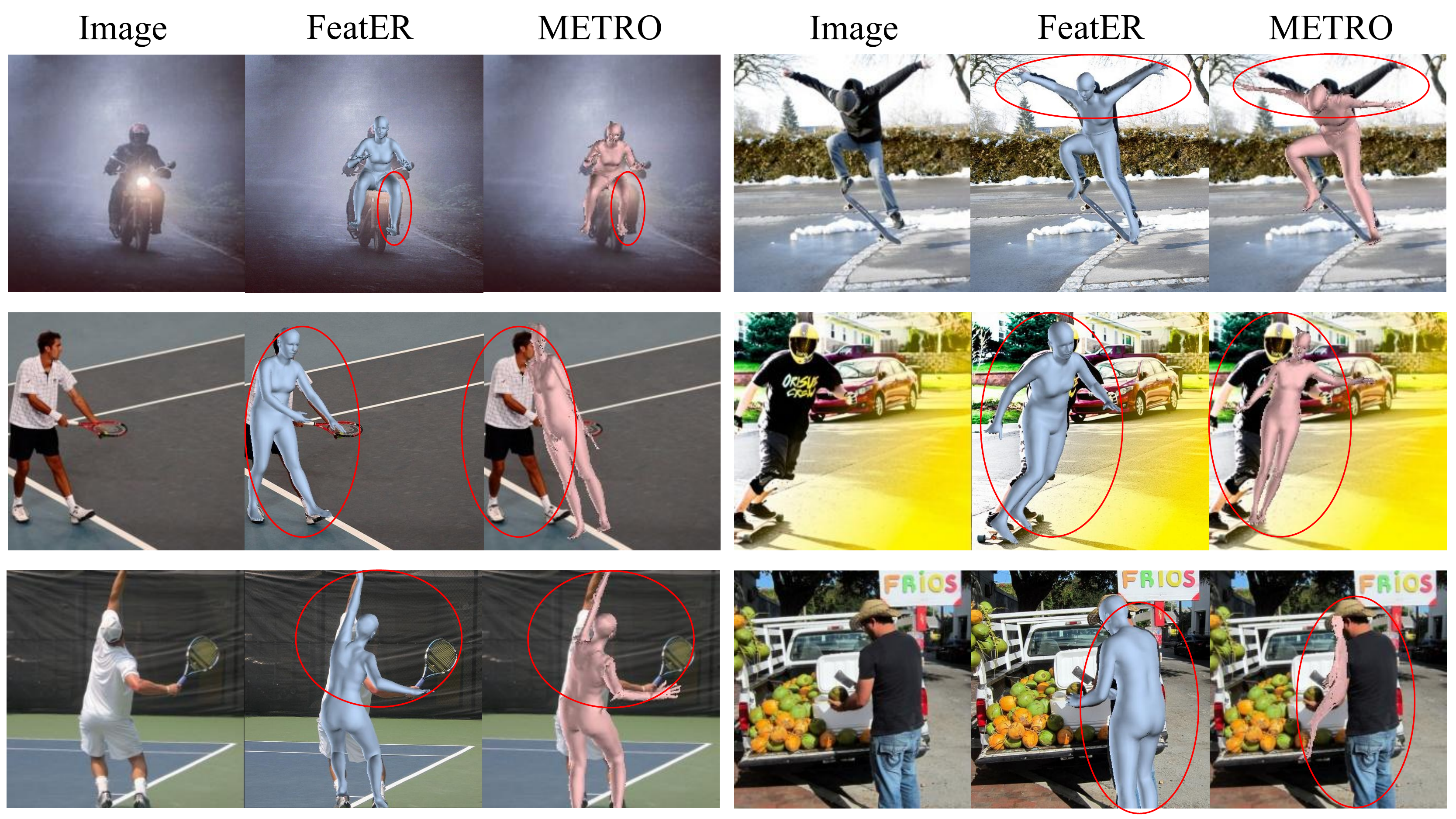}
  \caption{Qualitative comparison with the state-of-the-art HMR method METRO \cite{lin2021metro}. Images are taken from the in-the-wild COCO \cite{lin2014mscoco} dataset. The red circles highlight locations where FeatER is more accurate than METRO.}
  \label{fig:supp_compare}
  \vspace{-5pt}
\end{figure}

When comparing with the state-of-the-art HMR method METRO \cite{lin2021metro}, FeatER clearly outperforms METRO with only 5\% of Params and 16\% of MACs on these in-the-wild images (taken from the COCO \cite{lin2014mscoco} dataset) as depicted in Fig. \ref{fig:supp_compare}, demonstrating the superiority (in terms of both accuracy and efficiency) of the proposed FeatER method for practical applications. 

\textbf{Inaccurate and Failure Cases}

\begin{figure}[htp]
\vspace{-5pt}
  \centering
  \includegraphics[width=0.98\linewidth]{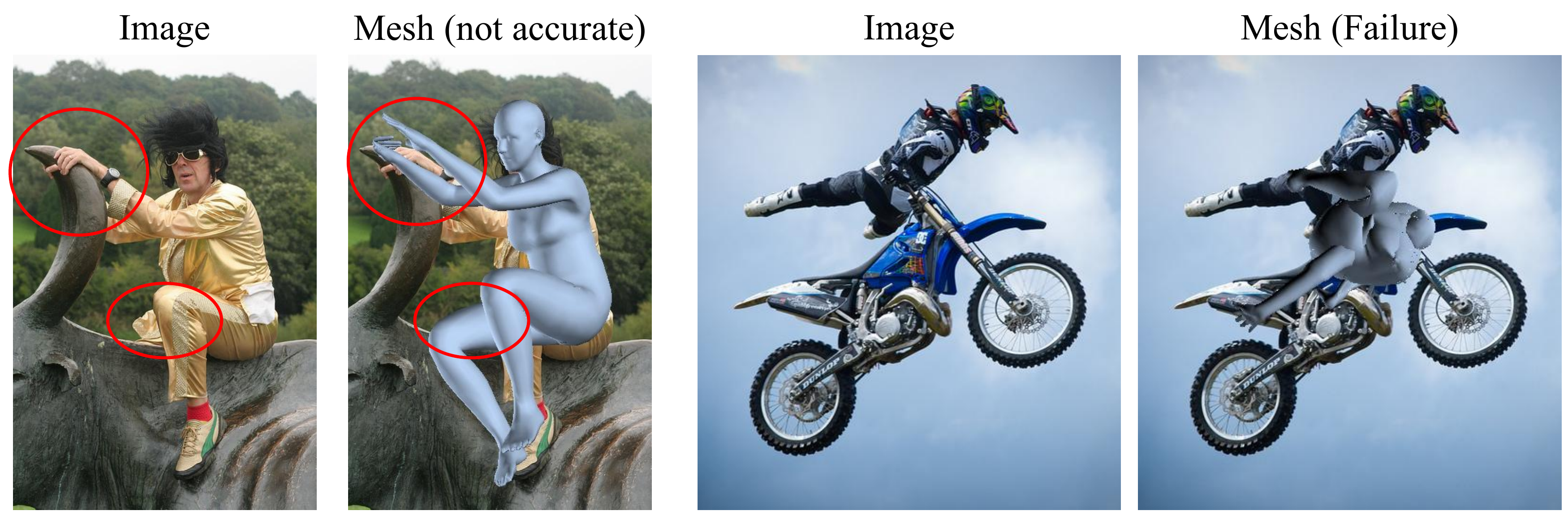}
  \caption{Left: Inaccurately estimated mesh due to heavy occlusion. Right: Failure estimated mesh due to complex human body articulation.}
  \label{fig:fail}
  \vspace{-5pt}
\end{figure}

Although FeatER can estimate human mesh quite well as demonstrated in Figs. \ref{fig:supp_easy} and \ref{fig:supp_hard}, there are still some inaccurate and failure cases. As presented in Fig. \ref{fig:fail} left, the red circle indicates the inaccurate mesh part due to heavy occlusion. The proposed Feature Map Reconstruction Module is not enough to tackle this issue with limited training data. For more complex human body articulation in Fig. \ref{fig:fail} right, FeatER fails to estimate accurate human mesh. How to further improve the generalization of FeatER to in-the-wild images would be our future work.

\section{Broader Impact and Limitations}
\label{Broader}
We believe that FeatER will help to highlight model efficiency for the HMR task. With significantly reduced computational and memory complexity than SOTA approaches, FeatER can still outperform them, which is better appreciated by real-world applications like VR/AR, virtual try-on, and AI coaching.

A potential limitation of FeatER is that it can not perform well in some specific scenarios such as crowded scenes. We leave this issue for future study.

{\small
\bibliographystyle{ieee_fullname}
\bibliography{egbib}

\begin{thebibliography}{10}\itemsep=-1pt

\bibitem{openpose}
Zhe Cao, Tomas Simon, Shih-En Wei, and Yaser Sheikh.
\newblock Realtime multi-person 2d pose estimation using part affinity fields.
\newblock In {\em Proceedings of the IEEE conference on computer vision and
  pattern recognition}, pages 7291--7299, 2017.

\bibitem{detr}
Nicolas Carion, Francisco Massa, Gabriel Synnaeve, Nicolas Usunier, Alexander
  Kirillov, and Sergey Zagoruyko.
\newblock End-to-end object detection with transformers.
\newblock In {\em European conference on computer vision}, pages 213--229.
  Springer, 2020.

\bibitem{chen2020dynamic}
Yinpeng Chen, Xiyang Dai, Mengchen Liu, Dongdong Chen, Lu Yuan, and Zicheng
  Liu.
\newblock Dynamic convolution: Attention over convolution kernels.
\newblock In {\em Proceedings of the IEEE/CVF Conference on Computer Vision and
  Pattern Recognition}, pages 11030--11039, 2020.

\bibitem{hpesurvey_chen}
Yucheng Chen, Yingli Tian, and Mingyi He.
\newblock Monocular human pose estimation: A survey of deep learning-based
  methods.
\newblock {\em Computer Vision and Image Understanding}, 192:102897, 2020.

\bibitem{cho2022FastMETRO}
Junhyeong Cho, Kim Youwang, and Tae-Hyun Oh.
\newblock Cross-attention of disentangled modalities for 3d human mesh recovery
  with transformers.
\newblock In {\em European Conference on Computer Vision (ECCV)}, 2022.

\bibitem{TCMR_Choi_2021}
Hongsuk Choi, Gyeongsik Moon, Ju~Yong Chang, and Kyoung~Mu Lee.
\newblock Beyond static features for temporally consistent 3d human pose and
  shape from a video.
\newblock In {\em Proceedings of the IEEE/CVF Conference on Computer Vision and
  Pattern Recognition (CVPR)}, pages 1964--1973, June 2021.

\bibitem{Choi_2020_ECCV_Pose2Mesh}
Hongsuk Choi, Gyeongsik Moon, and Kyoung~Mu Lee.
\newblock Pose2mesh: Graph convolutional network for 3d human pose and mesh
  recovery from a 2d human pose.
\newblock In {\em ECCV}, 2020.

\bibitem{Dosovitskiy2020ViT}
Alexey Dosovitskiy, Lucas Beyer, Alexander Kolesnikov, Dirk Weissenborn,
  Xiaohua Zhai, Thomas Unterthiner, Mostafa Dehghani, Matthias Minderer, Georg
  Heigold, Sylvain Gelly, Jakob Uszkoreit, and Neil Houlsby.
\newblock An image is worth 16x16 words: Transformers for image recognition at
  scale.
\newblock {\em ICLR}, 2021.

\bibitem{dsr2021}
Sai~Kumar Dwivedi, Nikos Athanasiou, Muhammed Kocabas, and Michael~J Black.
\newblock Learning to regress bodies from images using differentiable semantic
  rendering.
\newblock In {\em Proceedings of the IEEE/CVF International Conference on
  Computer Vision}, pages 11250--11259, 2021.

\bibitem{resnet}
Kaiming He, Xiangyu Zhang, Shaoqing Ren, and Jian Sun.
\newblock Deep residual learning for image recognition.
\newblock In {\em Proceedings of the IEEE conference on computer vision and
  pattern recognition}, pages 770--778, 2016.

\bibitem{TransReID}
Shuting He, Hao Luo, Pichao Wang, Fan Wang, Hao Li, and Wei Jiang.
\newblock Transreid: Transformer-based object re-identification.
\newblock In {\em Proceedings of the IEEE/CVF International Conference on
  Computer Vision (ICCV)}, pages 15013--15022, October 2021.

\bibitem{h36m_pami}
Catalin Ionescu, Dragos Papava, Vlad Olaru, and Cristian Sminchisescu.
\newblock Human3.6m: Large scale datasets and predictive methods for 3d human
  sensing in natural environments.
\newblock {\em IEEE Transactions on Pattern Analysis and Machine Intelligence},
  36(7):1325--1339, jul 2014.

\bibitem{OCHMR}
Rawal Khirodkar, Shashank Tripathi, and Kris Kitani.
\newblock Occluded human mesh recovery.
\newblock {\em Proceedings of the IEEE/CVF Conference on Computer Vision and
  Pattern Recognition (CVPR)}, 2022.

\bibitem{kingma2014adam}
Diederik~P Kingma and Jimmy Ba.
\newblock Adam: A method for stochastic optimization.
\newblock {\em arXiv preprint arXiv:1412.6980}, 2014.

\bibitem{kocabas2020vibe}
Muhammed Kocabas, Nikos Athanasiou, and Michael~J Black.
\newblock Vibe: Video inference for human body pose and shape estimation.
\newblock In {\em CVPR}, 2020.

\bibitem{Kolotouros2019SPIN}
Nikos Kolotouros, Georgios Pavlakos, Michael~J Black, and Kostas Daniilidis.
\newblock Learning to reconstruct 3d human pose and shape via model-fitting in
  the loop.
\newblock In {\em Proceedings of the IEEE/CVF International Conference on
  Computer Vision}, pages 2252--2261, 2019.

\bibitem{prohmr}
Nikos Kolotouros, Georgios Pavlakos, Dinesh Jayaraman, and Kostas Daniilidis.
\newblock Probabilistic modeling for human mesh recovery.
\newblock In {\em Proceedings of the IEEE/CVF International Conference on
  Computer Vision}, pages 11605--11614, 2021.

\bibitem{hybrik}
Jiefeng Li, Chao Xu, Zhicun Chen, Siyuan Bian, Lixin Yang, and Cewu Lu.
\newblock Hybrik: A hybrid analytical-neural inverse kinematics solution for 3d
  human pose and shape estimation.
\newblock In {\em Proceedings of the IEEE/CVF Conference on Computer Vision and
  Pattern Recognition}, pages 3383--3393, 2021.

\bibitem{PRTR}
Ke Li, Shijie Wang, Xiang Zhang, Yifan Xu, Weijian Xu, and Zhuowen Tu.
\newblock Pose recognition with cascade transformers.
\newblock In {\em Proceedings of the IEEE/CVF Conference on Computer Vision and
  Pattern Recognition}, pages 1944--1953, 2021.

\bibitem{li2021reid}
Ming Li, Jun Liu, Ce Zheng, Xinming Huang, and Ziming Zhang.
\newblock Exploiting multi-view part-wise correlation via an efficient
  transformer for vehicle re-identification.
\newblock {\em IEEE Transactions on Multimedia}, 2021.

\bibitem{li2022exploiting}
Wenhao Li, Hong Liu, Runwei Ding, Mengyuan Liu, Pichao Wang, and Wenming Yang.
\newblock Exploiting temporal contexts with strided transformer for 3d human
  pose estimation.
\newblock {\em IEEE Transactions on Multimedia}, 2022.

\bibitem{li2021mhformer}
Wenhao Li, Hong Liu, Hao Tang, Pichao Wang, and Luc Van~Gool.
\newblock Mhformer: Multi-hypothesis transformer for 3d human pose estimation.
\newblock {\em arXiv preprint arXiv:2111.12707}, 2021.

\bibitem{tokenpose2021}
Yanjie Li, Shoukui Zhang, Zhicheng Wang, Sen Yang, Wankou Yang, Shu-Tao Xia,
  and Erjin Zhou.
\newblock Tokenpose: Learning keypoint tokens for human pose estimation.
\newblock In {\em Proceedings of the IEEE/CVF International Conference on
  Computer Vision}, pages 11313--11322, 2021.

\bibitem{lin2021metro}
Kevin Lin, Lijuan Wang, and Zicheng Liu.
\newblock End-to-end human pose and mesh reconstruction with transformers.
\newblock In {\em Proceedings of the IEEE/CVF Conference on Computer Vision and
  Pattern Recognition}, pages 1954--1963, 2021.

\bibitem{lin2021_mesh_graphormer}
Kevin Lin, Lijuan Wang, and Zicheng Liu.
\newblock Mesh graphormer.
\newblock In {\em ICCV}, 2021.

\bibitem{lin2014mscoco}
Tsung-Yi Lin, Michael Maire, Serge Belongie, James Hays, Pietro Perona, Deva
  Ramanan, Piotr Doll{\'a}r, and C~Lawrence Zitnick.
\newblock Microsoft coco: Common objects in context.
\newblock In {\em European conference on computer vision}, pages 740--755.
  Springer, 2014.

\bibitem{liu2021group}
Ze Liu, Zheng Zhang, Yue Cao, Han Hu, and Xin Tong.
\newblock Group-free 3d object detection via transformers.
\newblock In {\em Proceedings of the IEEE/CVF International Conference on
  Computer Vision}, pages 2949--2958, 2021.

\bibitem{SMPL:2015}
Matthew Loper, Naureen Mahmood, Javier Romero, Gerard Pons-Moll, and Michael~J.
  Black.
\newblock {SMPL}: A skinned multi-person linear model.
\newblock {\em ACM TOG}, 2015.

\bibitem{shufflenetv2}
Ningning Ma, Xiangyu Zhang, Hai-Tao Zheng, and Jian Sun.
\newblock Shufflenet v2: Practical guidelines for efficient cnn architecture
  design.
\newblock In {\em Proceedings of the European conference on computer vision
  (ECCV)}, pages 116--131, 2018.

\bibitem{misra2021end}
Ishan Misra, Rohit Girdhar, and Armand Joulin.
\newblock An end-to-end transformer model for 3d object detection.
\newblock In {\em Proceedings of the IEEE/CVF International Conference on
  Computer Vision}, pages 2906--2917, 2021.

\bibitem{Moon_I2L_MeshNet}
Gyeongsik Moon and Kyoung~Mu Lee.
\newblock I2l-meshnet: Image-to-lixel prediction network for accurate 3d human
  pose and mesh estimation from a single rgb image.
\newblock In {\em ECCV}, 2020.

\bibitem{PyTorch}
Adam Paszke, Sam Gross, Soumith Chintala, Gregory Chanan, Edward Yang, Zachary
  DeVito, Zeming Lin, Alban Desmaison, Luca Antiga, and Adam Lerer.
\newblock Automatic differentiation in pytorch.
\newblock 2017.

\bibitem{vgg}
Karen Simonyan and Andrew Zisserman.
\newblock Very deep convolutional networks for large-scale image recognition.
\newblock {\em arXiv preprint arXiv:1409.1556}, 2014.

\bibitem{hrnet}
Ke Sun, Bin Xiao, Dong Liu, and Jingdong Wang.
\newblock Deep high-resolution representation learning for human pose
  estimation.
\newblock In {\em Proceedings of the IEEE/CVF Conference on Computer Vision and
  Pattern Recognition}, pages 5693--5703, 2019.

\bibitem{hmrsurvey}
Yating Tian, Hongwen Zhang, Yebin Liu, and Limin Wang.
\newblock Recovering 3d human mesh from monocular images: A survey.
\newblock {\em arXiv preprint arXiv:2203.01923}, 2022.

\bibitem{pw3d2018}
Timo von Marcard, Roberto Henschel, Michael Black, Bodo Rosenhahn, and Gerard
  Pons-Moll.
\newblock Recovering accurate 3d human pose in the wild using imus and a moving
  camera.
\newblock In {\em European Conference on Computer Vision (ECCV)}, sep 2018.

\bibitem{xiao2018simple}
Bin Xiao, Haiping Wu, and Yichen Wei.
\newblock Simple baselines for human pose estimation and tracking.
\newblock In {\em Proceedings of the European conference on computer vision
  (ECCV)}, pages 466--481, 2018.

\bibitem{xue2021transfer}
Fanglei Xue, Qiangchang Wang, and Guodong Guo.
\newblock Transfer: Learning relation-aware facial expression representations
  with transformers.
\newblock In {\em Proceedings of the IEEE/CVF International Conference on
  Computer Vision}, pages 3601--3610, 2021.

\bibitem{transpose2021}
Sen Yang, Zhibin Quan, Mu Nie, and Wankou Yang.
\newblock Transpose: Keypoint localization via transformer.
\newblock In {\em Proceedings of the IEEE/CVF International Conference on
  Computer Vision}, pages 11802--11812, 2021.

\bibitem{LiteHRNet}
Changqian Yu, Bin Xiao, Changxin Gao, Lu Yuan, Lei Zhang, Nong Sang, and
  Jingdong Wang.
\newblock Lite-hrnet: A lightweight high-resolution network.
\newblock In {\em CVPR}, 2021.

\bibitem{hrformer}
Yuhui Yuan, Rao Fu, Lang Huang, Weihong Lin, Chao Zhang, Xilin Chen, and
  Jingdong Wang.
\newblock Hrformer: High-resolution transformer for dense prediction.
\newblock {\em arXiv preprint arXiv:2110.09408}, 2021.

\bibitem{Thundr}
Mihai Zanfir, Andrei Zanfir, Eduard~Gabriel Bazavan, William~T Freeman, Rahul
  Sukthankar, and Cristian Sminchisescu.
\newblock Thundr: Transformer-based 3d human reconstruction with markers.
\newblock In {\em Proceedings of the IEEE/CVF International Conference on
  Computer Vision}, pages 12971--12980, 2021.

\bibitem{TCFormer}
Wang Zeng, Sheng Jin, Wentao Liu, Chen Qian, Ping Luo, Wanli Ouyang, and
  Xiaogang Wang.
\newblock Not all tokens are equal: Human-centric visual analysis via token
  clustering transformer.
\newblock In {\em Proceedings of the IEEE/CVF Conference on Computer Vision and
  Pattern Recognition}, pages 11101--11111, 2022.

\bibitem{darkpose_2020}
Feng Zhang, Xiatian Zhu, Hanbin Dai, Mao Ye, and Ce Zhu.
\newblock Distribution-aware coordinate representation for human pose
  estimation.
\newblock In {\em IEEE/CVF Conference on Computer Vision and Pattern
  Recognition (CVPR)}, June 2020.

\bibitem{pymaf2021}
Hongwen Zhang, Yating Tian, Xinchi Zhou, Wanli Ouyang, Yebin Liu, Limin Wang,
  and Zhenan Sun.
\newblock Pymaf: 3d human pose and shape regression with pyramidal mesh
  alignment feedback loop.
\newblock In {\em Proceedings of the IEEE International Conference on Computer
  Vision}, 2021.

\bibitem{shufflenetv1}
Xiangyu Zhang, Xinyu Zhou, Mengxiao Lin, and Jian Sun.
\newblock Shufflenet: An extremely efficient convolutional neural network for
  mobile devices.
\newblock In {\em Proceedings of the IEEE conference on computer vision and
  pattern recognition}, pages 6848--6856, 2018.

\bibitem{gtrs}
Ce Zheng, Matias Mendieta, Pu Wang, Aidong Lu, and Chen Chen.
\newblock A lightweight graph transformer network for human mesh reconstruction
  from 2d human pose.
\newblock {\em ACM multimedia}, 2022.

\bibitem{Poseformer_2021_ICCV}
Ce Zheng, Sijie Zhu, Matias Mendieta, Taojiannan Yang, Chen Chen, and Zhengming
  Ding.
\newblock 3d human pose estimation with spatial and temporal transformers.
\newblock In {\em Proceedings of the IEEE/CVF International Conference on
  Computer Vision (ICCV)}, pages 11656--11665, October 2021.

\end{thebibliography}
}

\end{document}